\documentclass[12pt]{article}
\usepackage{amsmath,amsfonts,amsthm,mathrsfs}
\usepackage{url}
\usepackage[page,title,titletoc,header]{appendix}
\usepackage{color}
\usepackage{fancyhdr}
\usepackage[top=2.8cm, bottom=2.8cm, left=3.4cm,
right=3.4cm]{geometry}
\usepackage{graphicx}
\newtheorem{example}{Example}
\newtheorem{theorem}{Theorem}
\newtheorem{assumption}{Assumption}
\newtheorem{lemma}{Lemma}
\newtheorem{proposition}{Proposition}
\newtheorem{remark}{Remark}
\newtheorem{corollary}{Corollary}
\newtheorem{definition}{Definition}

\allowdisplaybreaks[4]
\numberwithin{equation}{section}

\def\dsum{\displaystyle\sum }
\def\dsup{\displaystyle\sup }

\def\dint{\displaystyle\int }
\def\diint{\displaystyle\iint }
\def\begeqn{\begin{equation}}
\def\endeqn{\end{equation}}
\def\begth{\begin{theorem}}
\def\endth{\end{theorem}}
\def\begprop{\begin{proposition}}
\def\endprop{\end{proposition}}
\def\begcor{\begin{corollary}}
\def\endcor{\end{corollary}}
\def\begdef{\begin{definition}}
\def\enddef{\end{definition}}
\def\beglemm{\begin{lemma}}
\def\endlemm{\end{lemma}}
\def\begexm{\begin{example}}
\def\endexm{\end{example}}
\def\begrem{\begin{remark}}
\def\endrem{\end{remark}}
\def\begassum{\begin{assumption}}
\def\endassum{\end{assumption}}
\def\beg{\begin}
\def\ga{\alpha}

\def\gb{\beta}

\def\gga{{\gamma}}

\def\gd{\delta}

\def\gep{\varepsilon}

\def\gth{\theta}

\def\gk{\kappa}
\def\gl{\lambda}

\def\gs{\sigma}

\def\go{\omega}

\def\bz{{\bf z}}

\def\O{\mathcal{O}}

\def\N{\mathbb{N}}
\def\R{\mathbb{R}}

\def\X{\mathcal{X}}
\def\Y{\mathcal{Y}}
\def\Z{\mathcal{Z}}
\def\E{\mathcal{E}}

\def\L{\mathcal{L}}
\def\J{\mathcal{J}}
\def\K{\mathcal{K}}

\def\I{\mathcal{I}}
\def\H{\mathcal{H}}
\def\A{\hat{\mathcal{A}}}
\def\B{\hat{\mathcal{B}}}

\def\EX{{\mathbb{E}}}

\def\beg{\begin}
\def\Tr{\hbox{Tr}}

\def\gl{\lambda}
\def\R{\mathbb{R}}

\def\cR{{\mathcal{R}}}

\def\whL{\widehat{L}}

\def\wtf{\widetilde{f}}

\def\wtL{\widetilde{L}}
\def\wtS{\widetilde{S}}

\def\hx{\hat{x}}

\def\wt{\widetilde}

\def\EP{{\mathbb{P}}}

\begin{document}
\title{Online Pairwise Learning Algorithms with Kernels}
\author{Yiming Ying$^\dag$ and Ding-Xuan Zhou$^\ddag$\\
\\
$^\dag$Department of Mathematics and Statistics\\
State University of New York at Albany, Albany, NY 12222, USA\\
$^\ddag$Department of Mathematics, City University of Hong Kong\\
Tat Chee Avenue, Kowloon, Hong Kong, China}

\maketitle

\begin{abstract} Pairwise learning usually refers to a learning task which involves a loss function
depending on pairs of examples, among which most notable ones
include ranking, metric learning and AUC maximization.   In this paper, we study an online algorithm for pairwise learning with a least-square loss function in an unconstrained setting of a reproducing kernel Hilbert space (RKHS), which we refer to  as the Online Pairwise  lEaRning Algorithm (OPERA). In contrast to existing works \cite{Kar,Wang} which require that the iterates are restricted to a bounded domain or the loss function is strongly-convex, OPERA is associated with a non-strongly convex objective function and learns the target function in an unconstrained RKHS.  Specifically, we
establish a general theorem which guarantees the almost surely convergence  for the last iterate of
OPERA without any assumptions on the underlying distribution. Explicit convergence rates are derived  under the condition of polynomially decaying step sizes.  We also establish an interesting property for a family of widely-used kernels in the setting of pairwise learning and illustrate the above convergence results using such kernels. Our methodology mainly depends on the characterization of RKHSs using its associated
integral operators and probability inequalities for random variables
with values in a Hilbert space.
\end{abstract}

\section{Introduction}

For any $T\in\N$, the input space $\X$ is a compact domain  of $\R^d$ and the output space $\Y\subseteq\R$.
In the standard problems of regression and classification \cite{CZ,Steinwart}, one considers learning from a set
of examples $\bz = \{z_i=(x_i,y_i) \in \X \times \Y: i=1,2,\ldots, T\}$ drawn independently and
identically (i.i.d) from an unknown
distribution $\rho$ on $\Z = \X\times \Y.$ Associated with a specific learning problem, typically a
univariate loss function $\ell(h,x,y)$ is used to measure the quality of a hypothesis function $h: \X \to \Y$.

This paper is motivated by the growing interest in an important family of learning problems which, for
simplicity, we refer to as {\em pairwise learning} problems. In
contrast to classical regression and classification, such learning
problems involve  pairwise loss functions, i.e. the loss function
depends on a pair of examples which can be expressed by
$\ell(f,(x,y), (x',y'))$ for a hypothesis function $f: \X\times \X
\to \R.$ Many machine learning tasks can be formulated as pairwise
learning problems. Such tasks include ranking
\cite{Agarwal,ranksvm,Clem,Joachims,Rejchel}, similarity and metric
learning \cite{Bellet,CGY,Chechik,Weinberger1,YL}, AUC maximization
\cite{Zhao}, and gradient learning \cite{MW,MZ}.  For instance,
the task of ranking is to learn a ranking function capable of
predicting an ordering of objects according to some attached
relevance information. It generally involves the use of a misranking
loss $\ell(f,(x,y), (x',y')) = \mathbb{I}_{\{(y-y')f(x,x')<0\}}$ or
its surrogate loss  $\ell(f,(x,y), (x',y')) = (1-(y-y')f(x,x'))^2,$
where $\mathbb{I}(\cdot)$ is the indicator function. The goal of
ranking is to find a ranking rule $f$ in a hypothesis space $\H$
from the available data that minimizes the expected misranking risk
\begeqn \cR(f) = \diint_{\Z\times \Z} \ell(f, (x,y),
(x',y'))d\rho(x,y)d\rho(x',y').\endeqn In this paper, we assume that
the hypothesis function $f: \X \times \X \to \R $ for pairwise
learning belongs to a {\em reproducing kernel Hilbert space} (RKHS)
defined on the product space $\X^2= \X \times \X.$ Specifically, let
$K: \X^2 \times \X^2 \to \R$ be a {\em Mercer kernel}, i.e. a
continuous, symmetric and positive semi-definite kernel, see e.g.
\cite{CZ,Steinwart}. According to \cite{Aron}, the RKHS ${\mathcal
H}_{K}$ associated with kernel $K$ is defined to be the completion
of the linear span of the set of functions $\{K_{(x,x')}(\cdot):=
K((x,x'), (\cdot,\cdot)): (x,x')\in \X^2\}$ with an inner product
satisfying the { reproducing property}, i.e., for any $x',x\in \X$
and $f\in {\mathcal H}_K$, $ \langle K_{(x,x')}, f\rangle_{K} =
f(x,x'). $

Recently, a large amount of work focuses on  pairwise learning
algorithms in the batch setting in the sense that the algorithm uses
the training data $\bz$ at once. A general regularization scheme in
a RKHS ${\cal H}_K$ for pairwise learning can be formulated  as
\begeqn f_{\bz,\gl} =\arg \min_{f\in {\cal H}_K} \biggl\{{1 \over
T(T-1)} \sum_{i,j=1\atop i\neq j}^T \ell(f,(x_i,y_i), (x_j,y_j)) +
{\lambda \over 2 }\|f\|_K^2\biggr\}. \label{eq:batch}
\endeqn
where $\gl>0$ is a regularization parameter.  The above general
formulation was studied for ranking \cite{Agarwal,Rejchel} and
metric learning \cite{Bellet,CGY} under choices of different
pairwise kernels (see further discussions in Subsection
\ref{sec:pairwise-kernel}).  Their generalization analysis was
established using the concept of algorithmic stability
\cite{Agarwal}, robustness \cite{Bellet} or  U-statistics and
U-process \cite{CGY,Clem,Rejchel}. However, there is relatively
little work related to online algorithms for pairwise learning,
despite of its potential capability of dealing with large datasets.
Until most recently, \cite{Wang} established the first
generalization analysis of online learning methods for pairwise
learning in the linear case. In particular, they showed online to
batch conversion bounds hold true which are similar to those in the
univariate loss function case \cite{Cesa}.

In this paper, we study an Online Pairwise  lEaRning
Algorithms (OPERA) with a least-square loss function in a
reproducing kernel Hilbert space (RKHS). In particular,  a general convergence theorem
is established which guarantees the almost surely convergence of the last iterate of OPERA. Explicit convergence rates  are derived under the condition of polynomially decaying step sizes.   In contrast to existing works \cite{Kar,Wang} which require that the iterates are restricted to a bounded domain or the loss function is strongly-convex, OPERA is associated with a non-strongly convex objective function and learns the target function in an unconstrained RKHS (see more discussions in Section \ref{sec:related-work}).  Our novel methodology mainly depends on the characterization of RKHSs using
the associated integral operators and probability inequalities for random variables with values in the Hilbert space of Hilbert-Schmidt
operators.

The paper is organized as follows. Section \ref{sec:main-result} introduces OPERA and presents main results together with particular examples of specific pairwise kernels. Section \ref{sec:related-work} discusses the related work. Section
\ref{sec:technical-result} presents novel error decomposition for analyzing OPERA and establishes the
associated technical estimates.  The main results are proved in Section \ref{sec:proof-of-result}.  The paper concludes in Section
\ref{sec:conclusion}.  The proofs for technical lemmas are postponed to the Appendix.

\section{Main Results}\label{sec:main-result}

In this section, we introduce an online pairwise learning algorithm associated with
the least-square loss $\ell(f,(x,y),(x',y')) = (f(x,x')- y+y')^2$ in a reproducing
kernel Hilbert space $\H_K,$ and state our main results. In particular, denote the
true risk, for any function $f: \X\times
\X\to \R$, by $$\E(f) = \diint_{\Z\times \Z} (f(x,x')-y+y')^2d\rho(x,y)d\rho(x',y').$$
Denote by $\wtf_\rho$ the minimizer of the functional
$\E(\cdot)$ among all measurable functions. It is easy to see that $\wtf_\rho$ can be
represented by the difference of two standard regression
functions, i.e.
\begeqn\label{eq:regression-func}\wtf_\rho(x,x') = \dint_\X y d\rho(y|x) - \dint_\X y
d\rho(y|x') = f_\rho(x) - f_\rho(x').\endeqn
For this reason, throughout this paper we refer to $\wtf_\rho$ as the {\em pairwise regression function.} Denote by
$L^2_{\rho}(\X^2)$  the space of square
integrable functions on the domain $\X\times \X$, i.e. $$L^2_{\rho} (\X^2)=
\left\{f: \X\times \X \to \R:
\|f\|_\rho = \Big(\iint_{\X\times \X} |f(x,x')|^2
d\rho_\X(x)d\rho_\X(x')\Big)^{1/2}<\infty\right\},$$
where $\rho_\X$ is the marginal distribution of $\rho$ over $\X.$  Similar to the
standard least-square regression problem (see e.g. \cite{CS}), the
following property holds true $$ \E(f) - \E(\wtf_\rho) = \|f-\wtf_\rho\|^2_\rho.$$
In this paper, we study the following online pairwise learning algorithm
which aims to learn the pairwise regression function $\wtf_\rho$ from data.

\begdef Given the i.i.d. generated training data $\bz =
\{z_i=(x_i,y_i): i=1,2,\ldots,T\}$, the Online Pairwise lEaRning
Algorithm (OPERA) is given by $f_1=f_2 =0$ and, for $2\le t\le T$,
\begin{equation}\label{eq:algorithm} f_{t+1} = f_t - {\gga_t\over
t-1} \sum_{j=1}^{t-1}(f_t(x_t,x_j) - y_t+y_j) K_{(x_t,x_j)},
\end{equation} where $\{\gga_t>0: t\in \N\}$ is usually referred to
as the sequence of step sizes.
\enddef
OPERA is similar to the online projected gradient descent algorithm in \cite{Kar,Wang}, i.e.,  $f_0=0$ and $\eta = {R^2\over T}$, and, for $1\le t\le T$,
\begin{equation}\label{eq:opgd} f_{t} = \hbox{Proj}_{\mathcal{B}_R} \bigl[f_{t-1} - {\eta\over
t-1} \sum_{j=1}^{t-1}(f_t(x_t,x_j) - y_t+y_j) K_{(x_t,x_j)})\bigr],
\end{equation}
where $\hbox{Proj}_{\mathcal{B}_R}(\cdot)$ denotes the projection to a prescribed ball $\mathcal{B}_R= \{\|f\|_K\le R: f\in \H_K\}$ with radius $R$.  In contrast, OPERA does not have this additional projection step and is implemented in the unconstrained setting.

The sequence $\{f_t: t=1,2,\ldots,T+1\}$ is usually referred to as
the {\em learning sequence} generated by OPERA. We call the above
algorithm OPERA  an online learning algorithm in the sense that  it
only needs a sequential access to the training data. Specifically,
let $\bz^t = \{z_1,z_2,\ldots,z_t\}$ and at each time step $t+1$,
OPERA presumes a hypothesis $f_t\in \H_K$ upon which a new data
$z_t$ is revealed. The quality of the pairwise function $f_{t}$ is
estimated on the local  empirical error:
\begeqn\label{eq:local-loss} \widehat{\E}^t (f_t) = {1\over
2(t-1)}\sum_{j=1}^{t-1} (f_t(x_t,x_j)-y_t+y_j)^2.\endeqn The next
iterate $f_{t+1}$ given by equation (\ref{eq:algorithm}) is exactly
obtained by performing a gradient descent step from the current
iterate $f_t$ based on the gradient of the local empirical error,
which is  given by
$$  \nabla \widehat{\E}^t(f)|_{f= f_t} ={1\over t-1} \sum_{j =1}^{t-1} (f_t(x_t,x_j)-y_t+y_j) K_{(x_t,x_j)}.$$
Here, $\nabla \widehat{\E}^t (\cdot)$ denotes the functional
gradient of the functional $\widehat{\E}^t$ in the RKHS $\H_K.$

Now denote $\gk: = \dsup_{x,x\in \X} \sqrt{K((x,x'),(x,x'))},$ and
throughout the paper we assume that $|y|\le M$ almost surely for
some $M>0$. In addition, we introduce the notion of ${\mathcal
K}$-functional \cite{BL} in approximation theory as \begeqn
{\mathcal K}(s, \wtf_{\rho}):=\inf_{f\in {\mathcal
H}_K}\{\|f-\wtf_\rho\|_\rho+s\|f\|_K\}, \quad s>0.\label{kfcn}
\endeqn
We can establish the following general theorem about the convergence
of the last iterate $f_{T+1}$ generated by OPERA.

\begth\label{thm:1} Let $\gga_t = {1\over \mu}t^{-\gth}$ for any
$t\in \N$ with some $\gth \in ({1\over 2},1)$ and $\mu\ge \gk^2,$
and $\{f_t: t=1,\ldots, T+1\}$ be given by OPERA
(\ref{eq:algorithm}). For any $0<\gd<1$, we have with probability $1-\gd$  \begeqn\label{eq:converg}
\|f_{T+1} -\wtf_\rho\|_\rho  \le
\mathcal{K}\bigl(\sqrt{6\mu}(1+\gk)T^{-{1-\gth\over 2}},
\wtf_\rho\bigr)+  C_{\gth,\gk} \,
T^{-\min({\gth}-{1\over 2},{1-\gth\over 2})}\log T \log ({8T/\gd}), \endeqn where  $C_{\gth,\gk}$  depends on $\gk, \gth$ but independent of $T$ (see its explicit form in the proof).
\endth

Recall the well-known result (e.g. \cite{BL,YP}) that $$\lim_{s\to
0+} \K(s,\wtf_\rho) = \inf_{f\in\H_K}\|f-\wtf_\rho\|_\rho.$$ Then,
assuming $\gth\in (1/2,1)$ and letting $T\to \infty$ in inequality
(\ref{eq:converg}), we can prove the following corollary.
\beg{corollary}\label{cor:1} If $\gga_t = {1\over \mu}t^{-\gth}$ for
any $t\in \N$ with $\gth \in ({1\over 2},1)$ and $\mu\ge \gk^2$, and
$\{f_t: t=1,\ldots, T+1\}$ be given by OPERA (\ref{eq:algorithm}).
Then, $\|f_{T+1} - \wtf_\rho\|_\rho$ converges to
$\inf_{f\in\H_K}\|f-\wtf_\rho\|_\rho$ almost surely.
\end{corollary}
Let us discuss the implication of the above corollary.  Recall that
a kernel is universal if its associates RKHS is dense in the space
of continuous functions on $\X\times \X$ under the uniform norm.
Typical examples of universal kernels \cite{MXZ,Steinwart}
include the Gaussian kernel $K((x^1,x^2),(\hx^1,\hx^2)) =
\exp(-{\|(x^1,x^2)-(\hx^1,\hx^2)\|^2\over \gs})$ and the Laplace
kernel $K((x^1,x^2),(\hx^1,\hx^2)) =
\exp(-{\|(x^1,x^2)-(\hx^1,\hx^2)\|\over \gs}).$ In this case,
$\inf_{f\in\H_K}\|f-\wtf_\rho\|_\rho=0$, which equivalently implies
that, as $T\to \infty$, $\|f_{T+1}-\wtf_\rho\|_\rho \to 0$ almost surely.

We can derive explicit error rates under some regularity assumptions
on the pairwise regression function. The regularity of $\wtf_\rho$
can be typically measured by the integral operator $L_K: L^2_{\rho}
(\X^2) \to L^2_{\rho}(\X^2)$  defined by $$ L_K f = \iint_{\X\times
\X} f(x,x')K_{(x,x')}d\rho_\X(x)d\rho_\X(x').$$ Since $K$ is a
Mercer kernel, $L_K$ is compact and positive. Therefore, the
fractional power operator $L_K^\gb$ is well-defined for any $\gb>
0.$ In particular, we known from \cite{CS,CZ} that $L^{1/2}_K(L^2_\rho(\X^2)) = \H_K.$

\begth\label{thm:rates} Let $\{f_t: t=1,\ldots, T+1\}$ be given by
OPERA (\ref{eq:algorithm}). Suppose $\wtf_\rho\in L_K^{\gb}(L^2_\rho)$
with some $\gb>0$ and choose $\gga_t = {1\over \mu}
t^{-\min\bigr\{{2\gb+1\over 2\gb+2},{2\over 3}\bigr\}}$ with some
$\mu\ge \gk^2.$ Then, for any $0<\gd<1$ we have, with probability $1-\gd$, that  \begeqn\label{eq:rate-1}
\|f_{T+1}-\wtf_\rho\|_\rho \le C_{\gb,\gk} T^{-\min\bigl({\gb\over 2\gb+2},{1\over 6}\bigr)} \log T \log (8 T/\gd),\endeqn
where $C_{\gb,\gk}$ depends on $\gb,\gk$ and $\mu$ but independent of $T$ (see the explicit form in the proof).
\endth
The algorithm OPERA depends on  selecting an appropriate pairwise
kernel for a given learning task.  In the next subsection, we
consider a specific class of pairwise kernels and their associated
RKHSs which are induced by a kernel $G: \X\times \X \to \R.$

\subsection{Examples with specific pairwise kernels}\label{sec:pairwise-kernel}
Observe that the pairwise regression function $\wtf_\rho(x,x') =
f_\rho(x) - f_\rho(x')$, and hence  a natural motivation is to use a
pairwise function $f(x,x')=g(x) - g(x')$ to approximate the desired
function $\wtf_\rho,$ where $g\in \H_G$ with $G: \X\times \X \to \R$
being  a kernel.

Indeed, we can introduce a specific pairwise kernel $K$ such that
any function $f\in \H_K$ can be represented by as $f(x,x') =
g(x)-g(x')$ with $g\in \H_G.$  Specifically, given the univariate
kernel $G$, let the pairwise function $K: \X^2 \times \X^2
\to \R$ defined, for any $x^1, x^2, \hx^1, \hx^2 \in \X$, by
\begeqn\label{eq:MLPK}\beg{array}{ll} K((x^1,x^2),(\hx^1,\hx^2)) &=
G(x^1,\hx^1) + G(x^2,\hx^2) - G(x^1,\hx^2)
-G(x^2,\hx^1)\\
& = \langle G_{x^1} -G_{x^2}, G_{\hx^1} -
G_{\hx^2}\rangle_G.\end{array}\endeqn It can be easily verified that
$K$ defined by (\ref{eq:MLPK}) is positive semi-definite on
$\X^2\times \X^2,$ and thus  $K$ is a (pairwise) Mercer kernel on
$\X\times \X$ if $G$ is a Mercer kernel on $\X.$ The following
proposition characterizes the relationship between $\H_K$ and the
original RKHS $\H_G.$

\begprop \label{prop-1}Let $G: \X\times \X  \to \R$  be a Mercer
kernel and its associated pairwise kernel be induced by
(\ref{eq:MLPK}). Then, the following statements hold true:

\item (a) Assume the constant function $1_\X\in \H_G$ and let $\I_G = \hbox{span}\{1_\X\in \H_G\}$
containing all constant functions and ${\I}^\perp_G = \{g\in \H_G:
\langle g, 1_\X\rangle_G=0\}$ be the subspace orthogonal  to $\I_G$.
Then, the mapping $\Im: \I^\perp_G\to \H_K$ defined by
$\Im(g)(x^1,x^2) = g(x^1)-g(x^2)$ is a bijection with property
$\|\Im(g)\|_K = \|g\|_G.$

\item (b) If the constant function $1_\X\not\in \H_G$,  then the mapping $\Im: \H_G\to \H_K$ defined by
 $\Im(g)(x^1,x^2) = g(x^1)-g(x^2)$ is a bijection with property $\|\Im(g)\|_K = \|g\|_G.$

\endprop
Part (b) used the assumption $1_\X\not\in \H_G.$  Various kernels
induce RKHSs satisfying this assumption. For instance, the
homogeneous linear kernel $G(x,x') = x^\top x'$ and the Gaussian
kernel $G(x,x') = \exp(-{\|x-x'\|^2\over \gs})$ \cite{SHS} are such
kernels. However, in general the assumption in part (b) is not true,
and thus only part (a) holds true.

From the above proposition, we can rewrite OPERA
(\ref{eq:algorithm}) as $g_1=g_2=0$ and, for $ 2\le t \le T$,
\begin{equation}\label{eq:algorithm-2}
g_{t+1} = g_t - \gga_t \Big[{1\over t-1}
\sum_{j=1}^{t-1}(g_t(x_t)-g_t(x_j) - y_t+y_j) (G_{x_t} - G_{x_j})
\Big].\end{equation} The learning sequence $\{f_{t}:
t=1,2,\ldots,T+1\}$ of OPERA can be recovered by
\begeqn\label{eq:recover-alg}f_t(x^1,x^2) =\Im(g_{t})(x^1,x^2): =
g_t(x^1) -g_t(x^2), \qquad \quad \forall x^1,x^2\in \X.\endeqn
Denote $$L^2_{\rho} (\X)= \Bigl\{f: \X \to \R: \|f\|_\rho =
\bigl(\int_{\X} |f(x)|^2
d\rho_\X(x)\bigr)^{1/2}<\infty\Bigr\},$$ and, by applying Proposition \ref{prop-1},
we can see that the K-functional $\K$ defined by (\ref{kfcn}) is reduced to \begeqn\label{eq:K-G} {\mathcal K}_G(s, \wtf_{\rho}):=\left\{\begin{array}{ll}\inf_{g\in {\mathcal I}^\perp_G}\{\|\Im(g)-\wtf_\rho\|_\rho+s\|g\|_G\}, & \hbox{if}~ 1_\X \in \H_G \\
\inf_{g\in {\mathcal H}_G}\{\|\Im(g)-\wtf_\rho\|_\rho+s\|g\|_G\}, &  \hbox{otherwise}.
\end{array}\right.\endeqn
Equipped with the above notations,  we can obtain the following theorem. \begth \label{exmthm:1} Let $\gga_t = {t^{-\gth}\over
\gk^2}$ for any $t\in \N$ with $\gth \in ({1\over 2},1)$ and $\{g_t:
t=1,2,\ldots,T+1\}$ be given by algorithm (\ref{eq:algorithm-2}).
Then, the following statements hold true.
\item (a) Let the $\K$-functional associated with $\H_G$ be defined by (\ref{eq:K-G}). Then, for any $1<\gd<1$ we have, with probability $1-\gd$, that
 \begeqn\label{eq:converg-2} \|\Im(g_{T+1}) -
\wtf_\rho\|_\rho \le \mathcal{K}_G\bigl(\sqrt{6}\gk(1+\gk)T^{-{1-\gth \over 2}},
\wtf_\rho\bigr) + C_{\gth,\gk}\, T^{-\min({\gth}-{1\over 2},
{1-\gth\over 2})}\ln T.\endeqn
\item (b) Suppose $1_\X\not\in \H_G$ and $f_\rho\in L_G^{\gb}(L^2_\rho(\X))$ with some $0<\gb\le 1/2$ and
choose $\gga_t = {1\over \gk^2} t^{-{2\gb+1\over 2\gb+2}}.$ Then, for any $0<\gd<1$ we have, with probability $1-\gd$, that
\begeqn\label{eq:sepecial-rate-1} \|\Im(g_{T+1})-\wtf_\rho\|_\rho\bigr] \le  \widetilde{C}_{\gb,\gk}T^{-{\gb\over 2\gb+2}} \ln T.\endeqn
\endth
The above theorem implies the following result. Suppose that the original univariate kernel
$G$ is a Gaussian kernel in (\ref{eq:MLPK}).  choosing with $\gga_t =
{t^{-\gth}\over \gk^2}$ with $\gth\in (1/2,1)$ in (\ref{eq:algorithm-2}), by a similar argument to the proof for Corollary \ref{cor:1} we can have $\|\Im(g_{T+1})-\wtf_\rho\|_\rho \to 0$ almost surely as $T\to \infty.$  It remains a
question to us whether the assumption $1_\X\not\in \H_G$ in part (b)
of the above theorem can be removed  .

\section{Related work and Discussions}\label{sec:related-work}

In this section, we discuss the related work on pairwise learning
in the batch setting and stochastic online learning algorithms in
the univariate case.

Firstly, we briefly review  existing work  on pairwise learning,
among which most of them addressed the batch setting. In \cite{Rejchel}, the generalization analysis for the general formulation (\ref{eq:batch}) was conducted using
empirical process and U-statistics (see discussions in Example 3 there). Specifically, the author proved nice generalization
bounds for the excess risk of such estimators with rates faster than
$\O(1/\sqrt{T}),$ where $T$ is the sample number. In Section 5.2 of \cite{Agarwal}, the following
regularization formulation was studied for ranking: \begeqn\label{eq:Agarwal}
\min_{g\in {\cal H}_G} \biggl\{{2 \over T(T-1)} \sum_{i,j=1\atop
i<j}^T \psi(g(x_i)-g(x_j),y_i-y_j) + {\lambda \over 2
}\|g\|_G^2\biggr\},
\endeqn
where $\H_G$ denotes the RKHS on $\X$ with inner product
$\|\cdot\|_G$ and $\psi$ is a ranking loss function (see Definition
1 there).   This formulation can be regarded as a
special formulation of the general framework (\ref{eq:batch}) since,
by Proposition \ref{prop-1}, one can choose
$K((x^1,x^2),(\hx^1,\hx^2)) = \langle G_{x^1}-G_{x^2},
G_{\hx^1}-G_{\hx^2}\rangle_G$, and then, for any $f\in \H_K$, there
exists a $g\in \H_G$ such that $f(x_i,x_j) = g(x_i)- g(x_j)$ with
property $\|f\|_K = \|g\|_G.$  In contrast to the batch setting,
there is relatively little work on online algorithms for pairwise
learning.  Most recently, in \cite{Wang} and \cite{Kar}  online
to batch conversion bounds were nicely established for pairwise learning, which
shares the same spirit of \cite{Cesa} in the univariate case.  Specifically, Kar et~al. \cite{Kar} proved the following result. \footnote{The authors mainly focused on the linear case. However, the results there can be easily extended to the  kernelized case.}

\noindent{\bf Theorem A}$^{\cite{Kar}}$ {\em Let $f_1,f_2,\ldots,f_{T-1}$ be an ensemble of hypotheses from the space $\H$ generated by an online learning algorithm with a $B$-bounded loss function $\ell: \H\times \Z \times \Z\to [0, B]$ that guarantees a regret bound of $\Re_T$, i.e.  \begeqn\label{eq:regret-bound}\sum_{t=2}^{T} {1\over {t-1}}\sum_{\tau=1}^{t-1}\ell(f_{t-1},z_t,z_\tau) \le \inf_{f\in \H} \sum_{t=2}^{T} {1\over {t-1}}\sum_{\tau=1}^{t-1}\ell(f,z_t,z_\tau) + \Re_T.\endeqn
Then, for any $0<\gd<1$, we have with probability $1-\gd$,
$${1\over T-1} \sum_{t=2}^{T} \E_\ell(f_t)\le \inf_{f\in\H} \E_\ell(f) + {4\over T-1} \sum_{t=2}^T \mathcal{R}_{t-1}(\ell\circ \H) +{\Re_T\over T-1} + 6 B \sqrt{ \log {T\over \gd} \over T-1},$$
where, for any $f\in\H$, $\E_\ell(f) = \iint_{\Z\times\Z}  \ell(f, z,z')d\rho(z)d\rho(z'), $ and the Rademacher averages $ \mathcal{R}_T(\ell\circ \H)$ is defined as  $\mathcal{R}_{t-1}(\ell\circ \H)= \EX\bigl[ \sup_{h\in\H} {1\over t-1} \sum_{\tau=1}^{t-1} \gep_\tau \ell(h,z,z_\tau)\bigr]$ with the expectation being over $\gep_\tau,z,$ and $z_\tau.$}

For a fair comparison with our results, let the loss function $\ell(f,(x,y),(x',y')) = (f(x,x')- y+y')^2$ and the hypothesis space $\H $ be a bounded ball in an RKHS $\H_K$, i.e.  $\H =\mathcal{B}_R:= \{f\in\H_K:  \|f\|_K\le R\}$ with some $R>0.$ In this case, the constant $B$ in Theorem A is given by $B = (2M + \gk R)^2,$ and $\ell(\cdot, z,z')$ is Lipschitz continuous with constant $L = 2 M  + \gk R.$  By standard techniques to estimate the Rademacher averages, we can have $\mathcal{R}_t(\ell\circ \H) \le  \O ({R^2 \over \sqrt{t}})$ with $R$ sufficiently large.
Then, using an argument similar to the Section 5.3 of \cite{Wang} we know that the online projected gradient descent algorithm (\ref{eq:opgd}) enjoys the regret bound ${\Re}_T \le (2 M  + \gk R) R \sqrt{T}.$  Putting this regret bound with the above estimation for the Rademacher averages together, from Theorem A we get, with probability $1-\gd$, that
$$\beg{array}{ll}{1\over T-1} \sum_{t=2}^{T} \E(f_t) -\displaystyle\inf_{\|f\|_K\le R} \E(f) & \le \O\Bigl( {R^2\over T} \sum_{t=2}^T {1\over \sqrt{t} } +{R^2 \over \sqrt{T}}+R^2 \sqrt{ \log {T\over \gd} / (T-1)}\Bigr)\\
& \le\O\Bigl( R^2 \sqrt{ \log {T\over \gd} \over T}\Bigr).
\end{array}$$
Let $\overline{f}_T = {1\over T-1} \sum_{t=2}^{T} f_t$ and then we have $ \E(\overline{f}_T) \le {1\over T-1} \sum_{t=2}^{T} \E(f_t)$. Consequently,   $\E(\overline{f}_T)-\inf_{f\in\H} \E(f)\le \O\Bigl( R^2 \sqrt{ \log {T\over \gd} \over T}\Bigr). $
This estimation combined with the fact, for any $f$, that $\E(f)- \E(\wtf_\rho) = \|f-\wtf_\rho\|_\rho^2$ implies that
$$ \|\overline{f}_T-\wtf_\rho \|^2_\rho\le \inf_{\|f\|_K\le R} \|f-\wtf_\rho\|^2_\rho + \O\Bigl( R^2 \sqrt{ \log \Bigl({T\over \gd}\Bigr) / T}\Bigr).$$ The first term on the righthand side of the above inequality is known as approximation error.   Suppose the pairwise regression function $\wtf_\rho\in L_K^{\gb}(L^2_\rho)$ with some $0<\gb<1/2$. Then, we know from \cite{CZ,SZ} that $  \inf_{\|f\|_K\le R}\|f-\wtf_\rho\|^2_\rho \le R^{-{4\gb\over 1-2\gb}} \|L_K^{-\gb}\wtf_\rho\|^{2\over 1-2\gb}_\rho $, which implies that
$\|\overline{f}_T-\wtf_\rho \|^2_\rho\le \O\Bigl( R^{-{4\gb\over 1-2\gb}} \|L_K^{-\gb}\wtf_\rho\|^{2\over 1-2\gb}_\rho+ R^2 \sqrt{ \log \Bigl({T\over \gd}\Bigr) / T}\Bigr).$  Choosing $R = T^{1-2\gb\over 4} $ implies, with probability $1-\gd$, that
\begeqn\label{eq:rate-opgd2}\|\overline{f}_T-\wtf_\rho \|^2_\rho\le \O\Bigl(T^{-\gb} \bigl(\log\sqrt{T/\gd} + \|L_K^{-\gb}\wtf_\rho\|^{2\over 1-2\gb}_\rho\bigr)\Bigr).\endeqn
From Theorem \ref{thm:rates}, for $0<\gb<1/2$  the last iterate of OPERA has the convergence rate: \begeqn\label{eq:rate-opera2}\|f_T -\wtf_\rho \|^2_\rho\le \O\Bigl(T^{-{\gb\over 1+\gb}} (\log T \log (8 T/\gd))^2\Bigr).\endeqn
Comparing the rates in (\ref{eq:rate-opgd2}) and (\ref{eq:rate-opera2}), we can see that our rate (\ref{eq:rate-opera2}) for the last iterate of OPERA is suboptimal to that of the average of iterates generated by algorithm  (\ref{eq:opgd}). However, the online projected gradient descent algorithm (\ref{eq:opgd}) requires that all iterates are restricted to a prescribed ball with radius $R$, which leads to a challenging question on how to tune $R$ appropriately according to the real-data at hand.  In addition, the analysis techniques \cite{Kar,Wang} critically depend on the bounded-domain assumption and do not directly apply to the unconstrained setting here. OPERA is performed in the uncontrained setting and hence is parameter-free expect the choice of step sizes. Indeed, theorems in Section \ref{sec:main-result} show that choosing $\gga_t = \O(t^{-\gth})$ with $1/2<\gth<1$ always guarantees
that the last iterate of OPERA converges almost surely without   additional assumptions on the underlying distribution $\rho.$

Secondly, we discuss the related work on (stochastic) online
learning algorithms in the univariate case.   There is a large
amount work on (stochastic) online learning algorithms in the
univariate case \cite{Bottou,Cesa,Shamir,SY,YP,YZ} or under a more general
name called stochastic approximation
\cite{Bach,Nemirovski,Robbins}. The main idea is to use a
randomized gradient to replace the gradient
of the empirical loss, where the original idea dates back to the
work \cite{Robbins} in the 1950s. Most of approaches in
stochastic approximation assume the hypothesis space is of finite
dimensional and the gradient is bounded. In fact, when the
hypothesis space is of finite dimensional, a simple averaging scheme
for stochastic gradient descent \cite{Bach} can achieve the optimal
rate $\O({1\over T})$ under the assumption that the covariance
operator $\int_\X xx^\top d\rho_\X(x)$ is invertible. Stochastic
online learning with a least square loss in an infinite-dimensional
RKHS has been pioneered by \cite{SY} and the results were
established for general loss functions by \cite{YZ}, in which the
objective functions are all strongly convex.

OPERA (\ref{eq:algorithm}) shares a similar idea with the above
algorithms in the univariate case in the sense that, at each
iteration, it uses a computationally-cheap gradient estimator to
replace the true gradient. However, the objective function of OPERA
is not strongly convex and the hypothesis space $\H_K$ is not
bounded.  In particular, OPERA is more close to the online algorithm
in \cite{YP}, where the authors studied the following stochastic
gradient descent in a RKHS $\H_G$:
$$ \left\{\begin{array}{ll}  & g_1 = 0 ~ \hbox{ and }, \forall t\in
1,2,\ldots, T\\
 & g_{t+1} = g_t - \gga_t (g_t(x_t) - y_t)G_{x_t}.
 \end{array}\right.$$
The analysis in \cite{YP} heavily depends on the fact that the
randomized gradient $(g_t(x_t) - y_t)G_{x_t}$ is, conditionally on $\{z_1,z_2,\ldots,z_{t-1}\}$ , an unbiased
estimator of the true gradient $\iint_\X (g_t(x) - y)G_{x}
d\rho(x,y).$ However, the randomized gradient ${1\over t-1}
\sum_{j=1}^{t-1}(f_t(x_t,x_j) - y_t+y_j) K_{(x_t,x_j)}$ in OPERA
(\ref{eq:algorithm}) is not an unbiased estimator of the true
gradient $\iint_{\X\times \X}f_t(x,x') - y+y')
K_{(x,x')}d\rho(x,y)d\rho(x',y')$, even  conditionally on $\{z_1,z_2,\ldots,z_{t-1}\}.$ This introduces the main
difficulty in analyzing its convergence. Our new methodology relies
on the novel error decomposition presented in the next section. This
enable us to overcome this analysis difficulty by further employing
the characterization of RKHSs using the associated integral
operators and probability inequalities for random variables with
values in the Hilbert space of Hilbert-Schmidt operators.

\section{Error Decomposition and Technical Estimates}\label{sec:technical-result}
This section mainly presents an error decomposition for OPERA which
is critical to prove the main results in Section
\ref{sec:main-result}.

 To this end, we introduce some necessary notations. For any
$1\le j<t$, denote the linear operator
$$L_{(x_t,x_j)} = \langle \cdot, K_{(x_t,x_j)}\rangle_K
K_{(x_t,x_j)}: \H_K \to \H_K$$ by $L_{(x_t,x_j)}(g) = g(x_t,x_j)
K_{(x_t,x_j)}$ for any $ g\in \H_K,$ and let $\whL_{t} = {1\over
t-1} \sum_{j=1}^{t-1} L_{(x_t,x_j)}.$ In addition, define
$$S_{(z_t,z_j)} = (y_t-y_j)K_{(x_t,x_j)}, \hbox{ and }~ \hat{S}_{t}
={1\over t-1} \sum_{j=1}^{t-1}S_{(z_t,z_j)}.$$ We also define an
auxiliary operator $\wtL_t = \dint_\X \hat{L}_t d\rho(z_t)$, i.e.,
for any $f\in \H_K$
$$ \wtL_t (f) = {1\over t-1}\sum_{\ell=1}^{t-1} \dint_\X
f(x,x_\ell) K_{(x,x_\ell)} d\rho_\X(x).$$
Similarly, define $$\wtS_t = \dint_\X \hat{S}_t d\rho(z_t)
={1\over t-1} \sum_{\ell=1}^{t-1} \dint_{\X}
(f_\rho(x)-y_\ell)K_{(x,x_\ell)}d\rho_\X(x) .$$ In addition, let
$$\A^t = (\wtL_t- L_K) f_t -(\wtS_t - L_K\wtf_\rho), ~~
\B^t = (\hat{L}_t - \wtL_t )f_t -(\hat{S}_t -\wtS_t).$$

With these notations, for any $t\ge 2$ we can rewrite equality
(\ref{eq:algorithm}) as
$$f_{t+1} = f_t -  \gga_t (\hat{L}_t(f_t) - \hat{S}_t) = (I - \gga_t L_K) f_t - \gga_t (\hat{L}_t - L_K)(f_t) + \gga_t \hat{S}_t,$$
and \begeqn\label{eq:recursive-2}\beg{array}{ll} f_{t+1} -\wtf_\rho
&= (I-\gga_t L_K) (f_t-\wtf_\rho) - \gga_t (\hat{L}_t - L_K) f_t
+ \gga_t (\hat{S}_{t} - L_K \wtf_\rho) \\
& = (I-\gga_t L_K) (f_t-\wtf_\rho) - \gga_t \A^t - \gga_t \B^t.
\end{array}\endeqn
For any $t,j\in \N$ denote $\go^t_j (L_k) = \prod_{\ell=j}^t
(I-\gga_\ell L_K)$ for any $j\le t$ and we use the
conventional notation, for any $t\in \N$, $\go^t_{t+1} (L_k) =
I$ and $\sum_{\ell=t+1}^t\gga_\ell=0.$

Consequently, from the above equality we can derive, for any $t\ge
2$, that \begeqn\label{eq:recursive-3} f_{t+1} -\wtf_\rho = -
\go^t_2(L_K)(\wtf_\rho) - \dsum_{j=2}^t \gga_j\go^t_{j+1}(L_K)\A^j -
\dsum_{j=2}^t \gga_j\go^t_{j+1}(L_K)\B^j\endeqn The above error
decomposition is similar to the well-known ones in learning theory
in order to perform the error analysis for learning algorithms with
univariate loss functions, see e.g. \cite{Devito,SY,WYZ,YRC}. The
term $\go^t_2(L_K)(\wtf_\rho)$ is deterministic which is usually
referred to as {\em approximation error} and the other term, i.e.
$\sum_{j=2}^t \gga_j\go^t_{j+1}(L_K)\A^j+\sum_{j=2}^t
\gga_j\go^t_{j+1}(L_K)\B^j$, depends on the random samples which is
often called the {\em sample error}. Consequently, from the error
decomposition (\ref{eq:recursive-3}) we have
\begeqn\label{eq:estimate-bound-1}\begin{array}{ll}\|f_{t+1} -\wtf_\rho\|_\rho  & \le
\|\go^t_2(L_K)(\wtf_\rho)\|_\rho + \|\sum_{j=2}^t
\gga_j\go^t_{j+1}(L_K)\A^j\|_\rho \\ & + \|\sum_{j=2}^t
\gga_j\go^t_{j+1}(L_K)\B^j\|_\rho.\end{array}\endeqn In the following
subsections we estimate the terms on the right-hand side of
inequality (\ref{eq:estimate-bound-1}).

\subsection{Estimation of the sample error}
We now turn our attention to estimating the sample error, i.e. the
last two terms on the right-hand side of inequality
(\ref{eq:estimate-bound-1}). To this end, we first establish some
useful lemmas. The following lemma gives an upper-bound of the
learning sequence $\{f_t:t\in \N\}$ under the $\H_K$ norm, which is mainly inspired by a similar estimation in \cite{LZ} for bounding the iterates of online gradient descent algorithm in the univariate case.
\beglemm\label{lemm:bound} Let the learning sequence $\{f_t: t\in
\N\}$ be given by OPERA (\ref{eq:algorithm}) and assume, for any
$t\in \N$, that $\gga_t \gk^2\le 1.$ Then we have
\begeqn\label{eq:bound} \|f_{t}\|_K \le 2 M \sqrt{\dsum_{j=2}^{t-1}
\gga_j}, ~~~~\forall t\in \N.\endeqn
\endlemm
\beg{proof} For $t=1$ or $t=2$, by definition $f_1 = f_{2} =
0$ which certainly satisfy (\ref{eq:bound}). It suffices to
prove the case of $t\ge 2$ by induction. Recalling
equality (\ref{eq:algorithm}), we have
$$\beg{array}{ll}\|f_{t+1}\|_K^2 & = \|f_t\|^2_K  -{2\gga_t\over t-1} \dsum_{j=1}^{t-1} (f_t(x_t,x_j) - y_t+y_j)f_t(x_t,x_j)\\
& + {\gga_t^2\over (t-1)^2} \dsum_{j, j'=1}^{t-1} (f_t(x_t,x_j)-y_t+y_j)(f_t(x_t,x_{j'})-y_t+y_{j'}) K((x_t,x_j),(x_t,x_{j'}))\\
&\le \|f_t\|^2_K + {\gga_t^2 \gk^2\over t-1} \dsum_{j}^{t-1}
(f_t(x_t,x_j)-y_t+y_j)^2  \\
& -{2\gga_t\over t-1} \dsum_{j=1}^{t-1}
(f_t(x_t,x_j) - y_t+y_j)f_t(x_t,x_j).  \end{array}$$ Define a
univariate function $F_j$ by $F_j(s) = \gk^2\gga_t(s-y_t+y_j)^2 -
2(s-y_t+y_j)s.$ It is easy to see that $\sup_{s\in \R} F_j(s) =
{(y_t-y_j)^2\over 2-\gk^2\gga_t}\le (2M)^2$ since $\gga_t \gk^2\le
1$ and $|y_j|+|y_t| \le 2M.$ Therefore, from the above estimation we
can get, for $t\ge 2$, that
$$\beg{array}{ll} \|f_{t+1}\|^2_K & \le \|f_t\|_K^2
+ {\gga_t\over t-1} \dsum_{j=1}^{t-1}\sup_{j}F_j(s) \le \|f_t\|^2_K + (2 M)^2 \gga_t.
\end{array}$$
Combining the above inequality with the induction assumption that
$\|f_t\|_K \le 2 M \sqrt{\sum_{j=2}^{t-1}\gga_j}$ implies the
desired result. This completes the proof of the lemma.
\end{proof}

Denote the operator norm $\|\go_{j}^t(L_K)L^\gb_K\|_{\L(L^2_\rho)} =
\sup_{\|f\|_\rho\le 1} \|\go_{j}^t(L_K)L^\gb_K(f)\|_\rho.$ The
following technical lemma estimates the operator norm, which is
simply implied in the proof of Lemma 3 in \cite{YP}.
\beglemm\label{lemm:operator-norm} Let $\gb>0$ and $\gga_\ell\gk^2
\le 1 $ for any integer $ \ell\in [j,t].$ Then there holds
$$\|\go_{j}^t(L_K)L_K^{\gb} \|_{\L(L^2_\rho)}\le \bigl(\big(
{\gb\over e}\big)^{\gb}+\gk^{2\gb}\bigr) \min\Big\{1,
 \Big( \sum_{\ell=j}^t\gga_\ell\Big)^{-\gb}\Big\}.$$
\endlemm

The estimation of the sample error also relies on an important
characterization of $\H_K$ by the fractional operator $L_K^{1/2}$
(see Theorem 4 and Remark 3 in \cite{CS}). Specifically, for any
$f\in \H_K$ there exists $g\in L^2_\rho(\X^2)$ such that  $L^{1/2}_K
g = f$ with property $\|f\|_K=\|L_K^{1/2}g\|_K=\|g\|_\rho.$ With
this characterization of $\H_K,$ it is easy to see, for any $j<t$
and $f\in \H_K$, that
\begeqn\label{eq:hknorm-property}\beg{array}{ll}  \|\go^t_{j+1}(L_K)
f \|_\rho & =\|\go^t_{j+1}(L_K) L^{1/2}_K g \|_\rho   \le
\|\go^t_{j+1}(L_K) L^{1/2}_K\|_{\L(L^2_\rho)} \|g\|_\rho \\ & =
\|\go^t_{j+1}(L_K)
L^{1/2}_K\|_{\L(L^2_\rho)}\|f\|_K.\end{array}\endeqn

We also need the following probabilistic inequalities in a Hilbert space. The first one is the Bennett's inequality for random variables in Hilbert spaces, which can be easily derived from \cite[Theorem B 4]{SY}.
\beglemm\label{lem:Bennet} Let $\{\xi_i: i=1,2,\ldots,t\}$ be independent random variables in a Hilbert space $\H$ with norm $\|\cdot\|$. Suppose that almost surely $\|\xi_i\| \leq B $ and $ \EX \|\xi_i\|^2 \leq \gs^2<\infty$. Then, for any $0< \delta <1$, the following holds with probability at least $1-\delta$,
$$ \Big\|{1\over t} \sum_{i=1}^t [\xi_i - \EX \xi_i ]\Big\| \le {2 B \log {2\over \gd} \over t} + \gs \sqrt{\log {2\over \gd} \over t }$$

\endlemm

The second probabilistic inequality is the Pinelis-Bernstein inequality \cite[Proposition A.3]{YT} for martingale difference sequence in a Hilbert space, which is derived from \cite[Theorem 3.4]{P94}.

\begin{lemma}\label{Pinelis}
Let $\{S_k: k\in\N\}$ be a martingale difference sequence in a Hilbert space. Suppose that almost surely $\|S_k\| \leq B$ and $\sum_{k=1}^t \EX [\|S_k\|^2| S_1,\dots,S_{k-1}] \leq \gs_t^2$. Then, for any $0< \delta <1$, the following holds with probability at least $1-\delta$,
  $$\sup_{ 1\leq j\leq t} \left\|\sum_{k=1}^j S_k \right\| \leq 2 \left({B \over 3} + \gs_t\right) \log {2 \over \delta}.$$
\end{lemma}

We also need some facts on Hilbert-Schmidt operators on $\H_K,$ see
\cite{Devito,SZ}. Specifically, let $HS(\H_K)$ be the Hilbert space
of Hilbert-Schmidt operators on $\H_K$ with inner product $\langle
A, B \rangle_{HS} = \Tr(B^T A)$ for any $A, B \in HS(\H_K).$ Here
$\Tr$ denotes the trace of a linear operator. Indeed, the space
$HS(\H_K)$ is a subspace of the space of bounded linear operators on
$\H_K$, which is usually denoted by $(\L(\H_K),
\|\cdot\|_{\L(\H_K)})$ with the property, for any $A\in HS(\H_K)$,
that \begeqn\label{eq:norm-relation} \|A\|_{\L(\H_K)} \le
\|A\|_{HS}.\endeqn

With the above preparations, we are ready to estimate  the sample
error for algorithm
(\ref{eq:algorithm}) which, according to the error decomposition (\ref{eq:estimate-bound-1}),  consists of terms $\|\sum_{j=2}^t
\gga_j\go^t_{j+1}(L_K)\A^j\|_\rho$ and $\|\sum_{j=2}^t
\gga_j\go^t_{j+1}(L_K)\B^j\|_\rho.$ Let us start with the estimation of  $\|\sum_{j=2}^t
\gga_j\go^t_{j+1}(L_K)\A^j\|_\rho.$
\begth\label{thm:sample-A} Assume $\gga_t \gk^2\le 1$ for
any $t\in \N$ and let $\{f_t: t\in\N \}$ be given by equation
(\ref{eq:algorithm}). For any $t\ge 2$ and $0<\gd<1$, with probability $1-\gd$ there holds $$\|\sum_{j=2}^t
\gga_j\go^t_{j+1}(L_K)\A^j\|_\rho \le \bigl[12\gk(1+\gk)^2 M \log{4 t\over \gd} \bigr]
\sum_{j=2}^t {\gga_j(1+(\sum_{\ell=2}^{j-1}\gga_\ell)^{1/2}) \over
\sqrt{j}\bigl(1+\sum_{\ell=j+1}^t\gga_\ell\bigr)^{1/2}}.$$
\endth
\begin{proof} Write  $$ \sum_{j=2}^t
\gga_j\go^t_{j+1}(L_K)\A^j : = \sum_{j=2}^t
\gga_j\go^t_{j+1}(L_K)\A^j_1 + \sum_{j=2}^t
\gga_j\go^t_{j+1}(L_K)\A^j_2,$$ where $ \A^j_1 = (\wtL_j- L_K)
f_j$ and $\A^j_2 = -(\wtS_j - L_K\wtf_\rho).$ Hence,
\begeqn\label{eq:A-estimation}\beg{array}{ll} \|\dsum_{j=2}^t
\gga_j\go^t_{j+1}(L_K)\A^j\|_\rho &\le   \|\dsum_{j=2}^t
\gga_j\go^t_{j+1}(L_K)\A^j_1\|_\rho  \\ &  + \|\dsum_{j=2}^t
\gga_j\go^t_{j+1}(L_K)\A^j_2\|_\rho.\end{array}\endeqn For the first term on
the right-hand side of equation (\ref{eq:A-estimation}), we have
\begeqn\label{eq:A-estimation-1}\beg{array}{ll} &\|\dsum_{j=3}^t
\gga_j\go^t_{j+1}(L_K)\A^j_1\|_\rho  =\dsum_{j=3}^t \gga_j
\|\go^t_{j+1}(L_K)\A^j_1\|_\rho \\
& \le \dsum_{j=3}^t
\gga_j\|\go^t_{j+1}(L_K)L_K^{1/2}\|_{\L(L^2_\rho)}\;
\|\A^j_1\|_K\\
& \le \dsum_{j=3}^t
\gga_j\|\go^t_{j+1}(L_K)L_K^{1/2}\|_{\L(L^2_\rho)}\; \|\wtL_j-
L_K\|_{\L(\H_K)} \|f_j\|_K\\
& \le \dsum_{j=3}^t
\gga_j\|\go^t_{j+1}(L_K)L_K^{1/2}\|_{\L(L^2_\rho)}\; \|\wtL_j-
L_K\|_{HS} \|f_j\|_K,\end{array}\endeqn where the second 
inequality used (\ref{eq:hknorm-property}) and the last inequality
used (\ref{eq:norm-relation}).

Let the vector-valued random variable $\xi(x) = \int_\X \langle
\cdot, K_{(x',x)} \rangle_K K_{(x',x)}d\rho_\X(x')$. By following
the proof of Lemma 2 in \cite{Devito}, we have that $\|\langle
\cdot, K_{(x',x)} \rangle_K K_{(x',x)}\|_{HS}\le \gk^2.$ Hence,
$\|\xi\|_{HS} \le \int_\X \|\langle \cdot, K_{(x',x)} \rangle_K
K_{(x',x)}\|_{HS} d\rho_\X(x') \le \gk^2.$ Applying Lemma
\ref{lem:Bennet} with ${B} =\gs= \gk^2$ and $\H=
HS(\H_K)$, we have, with probability $1-{\gd\over t}$, that
\begeqn\label{eq:hilbert-rand-1}\begin{array}{ll} \|\wtL_j- L_K\|_{HS}
& =\big\|{1\over j-1}\dsum_{\ell=1}^{j-1}
\xi(x_\ell)-\EX(\xi)\big\|_{HS}\\ & \le {2\gk^2\log{2t \over \gd}\over {j-1}} + \gk^2 \sqrt{\log{2t \over \gd}\over j-1} \le {3\sqrt{2}\gk^2 \log{2t \over \gd}\over \sqrt{j}}. \end{array}\endeqn Applying Lemma
\ref{lemm:operator-norm} with $\gb=1/2$ implies, for any $2\le j\le
t$, that \begeqn\label{eq:operator-norm-1} \begin{array}{ll}
\|\go^t_{j+1}(L_K)L_K^{1/2}\|_{\L(L^2_\rho)} & \le \bigl(\big(
{1\over 2e}\big)^{1/2}+\gk\bigr) \min\Big\{1,
 \Big( \dsum_{\ell=j+1}^t\gga_\ell\Big)^{-1/2}\Big\}\\
& \le { \sqrt{2}\bigl( 1+\gk\bigr)/
\bigl(1+\sum_{\ell=j+1}^t\gga_\ell\bigr)^{1/2}},
 \end{array}\endeqn
where we used the conventional notation $\sum_{\ell=t+1}^t \gga_\ell
=0.$ Putting estimations (\ref{eq:hilbert-rand-1}),
(\ref{eq:operator-norm-1}) and inequality (\ref{eq:bound}) in Lemma
\ref{lemm:bound} back into (\ref{eq:A-estimation-1}), with probability $1-\gd$ there holds
\begeqn\label{eq:A-estimation-11}\|\sum_{j=3}^t
\gga_j\go^t_{j+1}(L_K)\A^j_1\|_\rho \le \bigl[12\gk^2(1+\gk)M\log{2t \over \gd}\bigr]
{\dsum_{j=3}^t {\gga_j(\sum_{\ell=2}^{j-1}\gga_\ell)^{1/2} \over
\sqrt{j}\bigl(1+\sum_{\ell=j+1}^t\gga_\ell\bigr)^{1/2}}}.\endeqn

For the term $ \|\dsum_{j=2}^t \gga_j\go^t_{j+1}(L_K)\A^j_2\|_\rho$,
we observe from (\ref{eq:hknorm-property}) again that
\begeqn\label{eq:A-estimation-2}\begin{array}{ll} \|\dsum_{j=2}^t
\gga_j\go^t_{j+1}(L_K)\A^j_2\|_\rho & \le \dsum_{j=2}^t
\gga_j\|\go^t_{j+1}(L_K)L_K^{1/2}\|_{\L(L^2_\rho)}\;
\|\A^j_2\|_K\\
& \le \dsum_{j=2}^t
\gga_j\|\go^t_{j+1}(L_K)L_K^{1/2}\|_{\L(L^2_\rho)}\; \| \wtS_j -
L_K\wtf_\rho\|_K. \end{array}\endeqn Let the vector-valued random
variable $\xi(z) = \int_\X (f_\rho(x')-y) K_{(x',x)}d\rho_\X(x')\in
\H_K$. Observe that $\|\xi\|_{K} \le \int_\X |f_\rho(x')-y|
\|K_{(x',x)}\|_{K} d\rho_\X(x') \le 2\gk M.$ Applying Lemma
\ref{lem:Bennet} with ${B} =\gs = 2\gk M $ and $\H= \H_K$, we have, with probability $1-{\gd\over t}$, that
$$\begin{array}{ll} \| \wtS_j - L_K\wtf_\rho\|_K &  = \|
{1\over j-1}\dsum_{\ell=1}^{j-1} \xi(z_\ell) - \EX(\xi)\|_K \\ & \le {4\gk M \log{2t\over \gd} \over {j-1}} + 2\gk M \sqrt{\log{2t\over \gd}\over j-1}\\
& \le {6\sqrt{2}\gk M  \log{2t\over \gd} \over \sqrt{j}}.
\end{array}$$
Putting the above estimation and inequality
(\ref{eq:operator-norm-1}) into (\ref{eq:A-estimation-2}) implies, with probability $1-\gd$,
that \begeqn\label{eq:A-estimation-21}\|\sum_{j=2}^t
\gga_j\go^t_{j+1}(L_K)\A^j_2\|_\rho\le \bigl[12\gk(1+\gk)M\log{2t\over \gd}\bigr]
\dsum_{j=2}^t { \gga_j\over
\sqrt{j}\bigl(1+\sum_{\ell=j+1}^t\gga_\ell\bigr)^{1/2}}.\endeqn
Combining inequalities (\ref{eq:A-estimation-11}) and
(\ref{eq:A-estimation-21}), we have, with probability $1-\gd$, that
$$\|\sum_{j=2}^t
\gga_j\go^t_{j+1}(L_K)\A^j\|_\rho \le
\bigl[12\gk(1+\gk)^2 M \log{4t\over \gd} \bigr] \sum_{j=2}^t
{\gga_j(1+(\sum_{\ell=2}^{j-1}\gga_\ell)^{1/2} )\over
\sqrt{j}\bigl(1+\sum_{\ell=j+1}^t\gga_\ell\bigr)^{1/2}}.$$
This completes the proof of the theorem.
\end{proof}

We move on to the estimation of the term $\|\sum_{j=2}^t
\gga_j\go^t_{j+1}(L_K)\B^j\|_\rho.$
\begth\label{thm:sample-B} Assume $\gga_t \gk^2\le 1$ for
any $t\in \N$ and let $\{f_t: t\in\N \}$ be given by equation
(\ref{eq:algorithm}). For any $t\ge 2$ and $0<\gd<1$, with probability $1-\gd$ there holds $$\|\sum_{j=2}^t
\gga_j\go^t_{j+1}(L_K)\B^j\|_\rho  \le {64 \over 3}\bigl(\gk(1+\gk)^2 M \log{2\over \gd}\bigr) \Bigl( \dsum_{j=2}^t   {\gga_j^2(1+\sum_{\ell=2}^{j-1} \gga_\ell) \over  1+\sum_{\ell=j+1}^{t} \gga_\ell}\Bigr)^{1\over 2}.$$
\endth
\begin{proof} Notice, from the recursive equality (\ref{eq:algorithm}), that  $f_j$ only
depends on samples $\{z_1,\ldots,z_{j-1}\}$ and $f_1 = f_2=0.$
Therefore, for any $j\ge 2$, there holds
\begeqn\label{eq:cancellation}\EX(\B^j | z_1,\ldots,z_{j-1})=0,\endeqn
which means that $\{\xi_j := \gga_j\go^t_{j+1}(L_K)\B^j: j=2,\ldots,t\}$ is a martingale difference sequence.
In the following, we will apply Lemma \ref{Pinelis} to estimate $\|\sum_{j=2}^t
\gga_j\go^t_{j+1}(L_K)\B^j\|_\rho.$ To this end, it remains to estimate
$B $ and $\gs_t^2.$

Recall that $\B^j = (\hat{L}_j - \wtL_j )f_j -(\hat{S}_j -\wtS_j).$ By  (\ref{eq:norm-relation}) and Lemma \ref{lemm:bound}, we have
$$\begin{array}{ll}\|\B_j\|_K  & \le \|\hat{L}_j - \wtL_j\|_{\L(\H_K)} \|f_j\|_K  + \|\hat{S}_j -\wtS_j\|_K\\
& \le \|\hat{L}_j - \wtL_j\|_{HS} \|f_j\|_K  + \|\hat{S}_j -\wtS_j\|_K\\
&\le 2 \gk^2 \|f_j\|_K  + 2\gk M \le 4 \gk^2 M \bigl(\dsum_{\ell=2}^{j-1} \gga_\ell\bigr)^{1\over 2} + 2\gk M.\end{array}$$
Consequently,  \begeqn\label{eq:B-estimation-1}\begin{array}{ll}
\|\go^t_{j+1}(L_K)\B_j\|_\rho
 & \le \|\go^t_{j+1}(L_K)L_K^{1/2}\|_{\L(L^2_\rho)}
\|\B^j_1\|_K \\ & \le   { \sqrt{2}( 1+\gk) \over
\bigl(1+\dsum_{\ell=j+1}^t\gga_\ell\bigr)^{1/2}}  \bigl(4 \gk^2 M \bigl(\sum_{\ell=2}^{j-1} \gga_\ell\bigr)^{1\over 2} + 2\gk M\bigr)\\
& \le 8 \gk(1+\gk)^2 M \Bigl( {1+\sum_{\ell=2}^{j-1} \gga_\ell \over  1+\sum_{\ell=j+1}^{t} \gga_\ell}\Bigr)^{1\over 2}. \end{array}
\endeqn
where the second inequality used Lemma (\ref{eq:operator-norm-1}).
From the above estimation, we have
$$\begin{array}{ll}   &\dsum_{j=2}^t \gga_j^2 \EX(\|\go^t_{j+1}(L_K)\B_j\|_\rho^2 |  z_1,\ldots,z_{j-1})\\ &  \le \gs_t^2 :=64 \gk^2(1+\gk)^4 M^2   \dsum_{j=2}^t   {\gga_j^2(1+\sum_{\ell=2}^{j-1} \gga_\ell) \over  1+\sum_{\ell=j+1}^{t} \gga_\ell},\end{array}$$
and
$$\begin{array}{ll}B & = \sup_{2\le j\le t} \gga_j\|\go^t_{j+1}(L_K)\B^j\|_\rho \le 8 \gk(1+\gk)^2 M \Bigl( \sup_{2\le j\le t}{\gga_j^2(1+\sum_{\ell=2}^{j-1} \gga_\ell) \over  1+\sum_{\ell=j+1}^{t} \gga_\ell}\Bigr)^{1\over 2} \\ &  \le 8 \gk(1+\gk)^2 M \Bigl( \dsum_{j=2}^t   {\gga_j^2(1+\sum_{\ell=2}^{j-1} \gga_\ell) \over  1+\sum_{\ell=j+1}^{t} \gga_\ell}\Bigr)^{1\over 2}.\end{array}$$
Applying Lemma \ref{Pinelis} yields that, with probability $1-\gd$,
$$\|\sum_{j=2}^t
\gga_j\go^t_{j+1}(L_K)\B^j\|_\rho  \le {64 \over 3}\bigl(\gk(1+\gk)^2 M \log{2\over \gd}\bigr) \Bigl( \dsum_{j=2}^t   {\gga_j^2(1+\sum_{\ell=2}^{j-1} \gga_\ell) \over  1+\sum_{\ell=j+1}^{t} \gga_\ell}\Bigr)^{1\over 2}.$$
This completes the proof of the theorem. \end{proof}

\subsection{Estimates of the approximation error}\label{subsec:appr}
Here, we establish some basic estimates for the deterministic
approximation error involving $\|\go^t_2(L_K){\wtf_\rho}\|_\rho.$ To
this end, we recall the notion of ${\mathcal K}$-functional
\cite{BL} in approximation theory, namely \begeqn {\mathcal K}(s,
f_{\rho}):=\inf_{f\in {\mathcal H}_K}\{\|f-f_\rho\|_\rho+s\|f\|_K\},
\quad s>0.\label{eq:kfcn}
\endeqn
We can estimate the quantity $\|\go^t_2(L_K)\wtf_\rho\|_\rho$ as follows.
\beglemm\label{preapplemm} Assume $\gga_t\gk^2 \le 1$ for each $t\in \N$. Then the following statements hold true.
\item (a) Let the $\K$-functional defined by (\ref{kfcn}). Then,
we have
\begeqn\label{eq:approx-K-functional}\|\go^t_2(L_K){\wtf_\rho}\|_\rho
\le \mathcal{K}\bigl(\sqrt{2}(1+\gk)\big(\dsum_{j=2}^t
\gga_j\big)^{-{1 \over 2}}, \wtf_\rho\bigr).\endeqn
\item (b) If $\wtf_\rho\in L^{\gb}_K({{\mathcal
L}^2_{\rho_X}})$ with some $\gb>0$ then \begeqn \|\go_{2}^t(L_K)
\wtf_\rho
\|_\rho \le { 2\biggl(\Big( {\gb\over e}\Big)^{\gb}+\gk^{2\gb}\biggr)}\|L^{-\gb}_K \wtf_\rho
\|_\rho\bigl(
\sum_{j=2}^t\gga_j \bigr)^{-\gb}. \label{gappbnd}\endeqn
\endlemm

\beg{proof} Part (a) is proved as follows. For any $f\in
{\mathcal H}_K$, from (\ref{eq:hknorm-property}) we have
\begeqn\label{eq:golkinterm}\beg{array}{ll}\|\go^t_2(L_K)\wtf_\rho\|_\rho
& \le \| f-\wtf_\rho\|_\rho+\|\go^t_2(L_K)f\|_\rho. \\
&=\|f-\wtf_\rho\|_\rho+\|\go^t_2(L_K)L_K^{{1 \over
2}}\|_{\L(L^2_\rho)}\|f\|_K.
\end{array}\endeqn Applying
Lemma \ref{lemm:operator-norm} with $\gb={1 \over 2},  j=2, $
implies that $\|\go^t_2(L_K)L_K^{{1 \over 2}}\|_{\L(L^2_\rho)}
\le
\sqrt{2}(1+\gk)\big(\dsum_{j=2}^t \gga_j\big)^{-{1 \over 2}}$.
Then,
substituting this into the right-hand side of
(\ref{eq:golkinterm})
yields that \begeqn \|\go^t_2(L_K)\wtf_\rho\|_\rho\le\inf_{f\in
{\mathcal H}_K}\Bigl\{\|f-\wtf_\rho\|_\rho
+\sqrt{2}(1+\gk)\big(\dsum_{j=2}^t \gga_j\big)^{-{1 \over 2}}
\|f\|_K\Bigr\} . \endeqn

Part (b) can be directly proved by applying Lemma
\ref{lemm:operator-norm} and the following observation
$$\|\go_{2}^t(L_K) \wtf_\rho \|_\rho \le
\|\go_{2}^t(L_K)L^{\gb}_K \|_{\L(L^2_\rho)}~\|L^{-\gb}_K
\wtf_\rho\|_\rho.$$
\end{proof}

\section{Proof of Main Results}\label{sec:proof-of-result}
In this section, we prove the results  presented in
Section \ref{sec:main-result}. Let us start with the proofs for Theorems \ref{thm:1} and \ref{thm:rates}. To this end, we need some technical lemmas.

\beglemm\label{lemm:A} Let $\gga_j= {j^{-\gth}\over \mu}$ for any
$j\in\N$ with $\gth\in ({1\over 2},1)$ and $\mu>0$. Then we have,
for any $t\ge 4$, that \begeqn\label{eq:A-sum-est}\sum_{j=2}^t
{\gga_j (1+\sum_{\ell=2}^{j-1}\gga_\ell) \over \sqrt{j}\bigl(1+
\sum_{\ell=j+1}^t \gga_\ell\bigr)^{1/2}} \le {C}_\gth
t^{-\min({\gth}-{1\over 2},{1-\gth\over 2})}\log t, \endeqn where $$
{C}_\gth =  \left\{\beg{array}{ll} {26
\max\bigl(\sqrt{\mu(1-\gth)})^{-1}, \sqrt{\mu(1-\gth)}\bigr)\over
\mu(1-\gth)|3\gth-2|} + \sqrt{5\over 2\mu}, &  \hbox{ if } \gth\neq
2/3\\ {20 \max\bigl(\sqrt{\mu(1-\gth)})^{-1},
\sqrt{\mu(1-\gth)}\bigr)\over \mu(1-\gth)}+ \sqrt{5\over 2\mu}, &
\hbox{ if } \gth=2/3.\end{array}\right.$$
\endlemm

\beglemm\label{lemm:B} Let $\gga_j= {j^{-\gth}\over \mu}$ for any
$j\in\N$ with $\gth\in (0,1)$. Then we have, for any $t\ge 4$, that
\begeqn\label{eq:B-sum-est}\Bigl(\sum_{j=2}^t {\gga_j^2
(1+(\sum_{\ell=2}^{j-1}\gga_\ell)^2) \over 1+ \sum_{\ell=j+1}^t
\gga_\ell}\Bigr)^{1/2} \le \widetilde{C}_\gth
t^{-\min({\gth}-{1\over 2},{1-\gth\over 2})}\log t, \endeqn where $
\widetilde{C}_\gth =\left\{\begin{array}{ll} \Bigl( {5\over 8\mu} +
{16\max((\mu(1-\gth))^{-1}, \mu(1-\gth))\over
\mu^2(1-\gth)|3\gth-2|}\Bigr)^{1/2}, &  \hbox{ if } \gth\neq
2/3 \\
\Bigl( {5\over 8\mu} + {16\max((\mu(1-\gth))^{-1}, \mu(1-\gth))\over
\mu^2 (1-\gth)} \Bigr)^{1/2}, & \hbox{ if }
\gth=2/3.\end{array}\right.$
\endlemm

The proofs for Lemma \ref{lemm:A} and Lemma \ref{lemm:B} are given in the Appendix. With the above lemmas, we are ready to
establish the main results stated in Section \ref{sec:main-result}.

\noindent{\bf Proof of Theorem \ref{thm:1}.} Applying
(\ref{eq:estimate-bound-1}) with $t=T$, we have
\begeqn\label{eq:estimate-bound-2}\begin{array}{ll}\|f_{T+1} -\wtf_\rho\|_\rho &\le
\|\go^T_2(L_K)(\wtf_\rho)\|_\rho + \|\sum_{j=2}^T
\gga_j\go^T_{j+1}(L_K)\A^j\|_\rho \\ & + \|\sum_{j=2}^T
\gga_j\go^T_{j+1}(L_K)\B^j\|_\rho.\end{array}\endeqn
By Theorem
\ref{thm:sample-A} and (\ref{eq:A-sum-est}), with probability $1-\gd$, there holds
\begeqn\label{eq:term2}\beg{array}{ll}\|\sum_{j=2}^T
\gga_j\go^T_{j+1}(L_K)\A^j\|_\rho & \le
12\gk(1+\gk)^2M \log{4T\over \gd} \dsum_{j=2}^T
{\gga_j(1+\sum_{\ell=2}^{j-1}\gga_\ell) \over
\sqrt{j}\bigl(1+\sum_{\ell=j+1}^t\gga_\ell\bigr)^{1/2}}\\
& \le 12 C_\gth\, \gk(1+\gk)^2M \,T^{-\min({\gth}-{1\over 2},{1-\gth\over 2})} \log T \log {4 T\over \gd} \end{array}\endeqn
From  Theorem
\ref{thm:sample-B} and (\ref{eq:B-sum-est}) we have, with probability $1-\gd$, that
\begeqn\label{eq:term3}\beg{array}{ll}\|\sum_{j=2}^T
\gga_j\go^T_{j+1}(L_K)\B^j\|_\rho &\le {64\over 3}\gk(1+\gk)^2M \Bigl(\sum_{j=2}^T {\gga_j^2
(1+(\sum_{\ell=2}^{j-1}\gga_\ell)^2) \over 1+ \sum_{\ell=j+1}^T
\gga_\ell}\Bigr)^{1/2}\\
& \le {64\widetilde{C}_\gth\over 3}\gk(1+\gk)^2M\,
T^{-\min({\gth}-{1\over 2},{1-\gth\over 2})}\log T  \log {2\over \gd}.
\end{array}\endeqn
Putting estimates (\ref{eq:estimate-bound-2}), (\ref{eq:term2}), and
(\ref{eq:term3}), with probability $1-2\gd$ there holds \begeqn\label{eq:gen-est}
\|f_{T+1} -\wtf_\rho\|_\rho  \le
\|\go^T_2(L_K)(\wtf_\rho)\|_\rho+  C_{\gth,\gk} \,
T^{-\min({\gth}-{1\over 2},{1-\gth\over 2})}\log T \log {4T \over \gd},\endeqn
where $C_{\gth,\gk} = 4(3C_\gth + {16\widetilde{C}_\gth\over 3}) \gk (1+ \gk)^2 M.$

In addition, by (\ref{eq:approx-K-functional}), we have
$$\|\go^T_2(L_K)(\wtf_\rho)\|_\rho \le
\mathcal{K}\bigl(\sqrt{2}(1+\gk)\big(\dsum_{j=2}^T \gga_j\big)^{-{1
\over 2}}, \wtf_\rho\bigr).$$ Notice that $\sum_{j=2}^T \gga_j =
{1\over \mu}\sum_{j=2}^T j^{-\gth}\ge
{(T+1)^{1-\gth}-2^{1-\gth}\over \mu(1-\gth)} \ge {(1-({2\over
3})^{1-\gth})(T+1)^{1-\gth} \over \mu(1-\gth)}\ge {T^{1-\gth}
\over 3\mu(1-\gth)} \ge {T^{1-\gth}\over 3\mu}.$ Consequently,
\begeqn\label{eq:term1}\|\go^T_2(L_K)(\wtf_\rho)\|_\rho \le
\mathcal{K}\bigl(\sqrt{6\mu}(1+\gk)T^{-{1-\gth\over 2}},
\wtf_\rho\bigr).\endeqn
Putting this back into (\ref{eq:gen-est}) implies the desired result. This completes the proof of the theorem. \hfill $\Box$

\noindent{\bf Proof of Corollary \ref{cor:1}.}   By the definition of the almost surely convergence, it suffices to prove, for any $\gep>0$, that
$$\lim_{t_0\to \infty}\EP\bigl(\sup_{t\ge t_0}[\|f_{t+1} -\wtf_\rho\|_\rho - \inf_{f\in \H_K}\|f -\wtf_\rho\|_\rho] \ge 2\gep\bigr)=0.$$
However, it is well-known that $\lim_{s\to +0} \K(s, \wtf_\rho) =
\inf_{f\in \H_K} \|f-\wtf_\rho\|_\rho$ (see e.g. Lemma 9 in \cite{YP}). This means that there exists $t_1\in\N$ such that, for any $t\ge t_1$, there holds $$ \mathcal{K}\bigl(\sqrt{6}\gk(1+\gk)t^{-{1-\gth\over 2}},
\wtf_\rho\bigr) -\inf_{f\in \H_K} \|f-\wtf_\rho\|_\rho \le {\gep}.$$
Let $\cR_{t} = \|f_{t+1} -\wtf_\rho\|_\rho - \mathcal{K}\bigl(\sqrt{6}\gk(1+\gk)t^{-{1-\gth\over 2}},
\wtf_\rho\bigr).$ The above estimation implies, for any $t_0\ge t_1$, that
\begeqn\label{eq:as-1}\begin{array}{ll}&\EP\bigl(\dsup_{t\ge t_0} [\|f_{t+1} -\wtf_\rho\|_\rho - \inf_{f\in \H_K}\|f -\wtf_\rho\|_\rho]\ge 2\gep\bigr) \\ & \le \EP\bigl(\dsup_{t\ge t_0} \cR_t \ge {\gep}\bigr)   \le \dsum_{t=t_0}^\infty \EP\bigl( \cR_t \ge {\gep}\bigr).\end{array}\endeqn
From Theorem 1,   we have, for any $1-\gd$, that  $$\EP\Bigl( \cR_t \ge  C_{\gth,\gk} \,
t^{-\min({\gth}-{1\over 2},{1-\gth\over 2})}\log t \log ({4t/\gd})\Bigr)\le \gd. $$
which is equivalent to
$$\EP\Bigl( \cR_t \ge  \gep \Bigr) \le  4 t \exp\Bigl(-{t^{\min(\gth-{1\over 2},{1-\gth\over 2})}\gep \over C_{\gth,\gk}\log t}\Bigr).$$
Putting this back into (\ref{eq:as-1}) implies that
 \begeqn\label{eq:as-2}\EP\bigl(\dsup_{t\ge t_0} [\|f_{t+1} -\wtf_\rho\|_\rho - \inf_{f\in \H_K}\|f -\wtf_\rho\|_\rho]\ge 2\gep\bigr) \le \dsum_{t=t_0}^\infty 4 t \exp\Bigl(-{t^{\min(\gth-{1\over 2},{1-\gth\over 2})}\gep \over C_{\gth,\gk}\log t}\Bigr).\endeqn
For any $1/2<\gth<1$ and $\gep>0$,  it is easy to see that $$\dsum_{t=2}^\infty 4 t \exp\bigl(-{t^{\min(\gth-{1\over 2},{1-\gth\over 2})}\gep \over C_{\gth,\gk}\log t}\bigr)<\infty.$$  Consequently,
$$\lim_{t_0\to \infty}\dsum_{t=t_0}^\infty 4 t \exp\Bigl(-{t^{\min(\gth-{1\over 2},{1-\gth\over 2})}\gep \over C_{\gth,\gk}\log t}\Bigr)=0.$$
Combining this with (\ref{eq:as-2}) implies, for any $\gep>0$, that
$$\lim_{t_0\to \infty}\EP\bigl(\sup_{t\ge t_0}[\|f_{t+1} -\wtf_\rho\|_\rho - \inf_{f\in \H_K}\|f -\wtf_\rho\|_\rho] \ge 2\gep\bigr)=0.$$
This completes the proof of the corollary.
 \hfill $\Box$

From Theorem \ref{thm:1} and the estimation (\ref{gappbnd}) for the
approximation error, we can derive the explicit error rates for
OPERA stated in Theorem \ref{thm:rates}.

\noindent{\bf Proof of Theorem \ref{thm:rates}.}  Applying
(\ref{gappbnd}) with $\gb>0$ and (\ref{eq:sum-poly}) with $\gga_\ell
= {1\over \mu}\ell^{-\gth}$,  $j=2$ and $ k=T$, we have that
$$ \begin{array}{ll} & \|\go_{2}^T(L_K)L_K^{\gb}
\|_{\L(L^2_\rho)}\le \bigl(\big(
{\gb\over e}\big)^{\gb}+\gk^{2\gb}\bigr)
 \Big( \sum_{\ell=2}^T\gga_\ell\Big)^{-\gb}   \le  \bigl(\big(
{\gb\over e}\big)^{\gb}+\gk^{2\gb}\bigr)
 \Big( \sum_{\ell=2}^T {1\over \mu} \ell^{-\gth}\Big)^{-\gb}\\
 & \le \bigl(\big(
{\gb\over e}\big)^{\gb}+\gk^{2\gb}\bigr)\gk^{2\gb} \mu^{\gb}
 \Big( \sum_{\ell=2}^T \ell^{-\gth}\Big)^{-\gb}\\
& \le \bigl(\big(
{\gb\over e}\big)^{\gb}+\gk^{2\gb}\bigr)\gk^{2\gb} (\mu(1-\gb))^{\gb}
 \Big( T^{1-\gth}-1\Big)^{-\gb}\\
  &\le \bigl[\bigl(\big(
{\gb\over e}\big)^{\gb}+\gk^{2\gb}\bigr)\gk^{2\gb} (\mu(1-\gb))^{\gb}
 (1-({1\over 2})^{1-\gth})^{-\gb}\bigr]T^{-\gb(1-\gth)} := D_{\gk,\gb} T^{-\gb(1-\gth)}
 \end{array}$$
Putting this estimation into Theorem \ref{thm:1} yields, with probability $1-\gd$, that
\begeqn\label{eq:rate-gen} \|f_{T+1}-\wtf_\rho\|_\rho
\le D_{\gk,\gb}  T^{-\gb(1-\gth)} +  C_{\gth,\gk} T^{-\min\{{\gth}-{1\over 2}, {1-\gth\over
2}\}}\log T \log (8T/\gd)).\endeqn Selecting $\gth= \min\bigr\{{2\gb+1\over 2\gb+2},{2\over 3}\bigr\}$ implies, for probability $1-\gd$, that
$$\|f_{T+1}-\wtf_\rho\|_\rho \le (D_{\gk,\gb}+ C_{\gth,\gk})  T^{-\min\bigl({\gb\over 2\gb+2},{1\over 6}\bigr)} \log T \log(8T/\gd).$$ This completes the proof of the
theorem. \hfill $\Box$

We now turn our attention to the special pairwise kernel
(\ref{eq:MLPK}) induced by a univariate kernel $G$. Let us
first prove Proposition \ref{prop-1} which describes the
relationship between the space $\H_K$ with the pairwise kernel $K$
and $\H_G$ with the univariate kernel $G$.

\noindent {\bf Proof of Proposition \ref{prop-1}.}  To prove (a),
for any $n\in \N$, $\{\ga_i: i=1,\ldots, n\}$ and
$\{(x_i^1,x^2_i)\in \X \times \X: i=1,\ldots,n\}$, let $g =
\sum_{i=1}^n  \ga_i (G_{x^1_i}- G_{x^2_i})\in \H_G.$ Indeed, it can
be further be verified that $g\in \I^\perp_G$ since $\langle g,
1_\X\rangle_G =\langle \sum_{i=1}^n  \ga_i (G_{x^1_i}- G_{x^2_i}),
1_\X\rangle_G = \sum_{i=1}^n  \ga_i (1_\X(x^1_i)- 1_\X(x^2_i))=0.$
Then, for any $x^1,x^2\in \X$,
$$\beg{array}{ll}\sum_{i=1}^n  \ga_i K_{(x^1_i,x^2_i)}(x^1,x^2) & = \sum_{i=1}^n  \ga_i (G_{x^1_i}(x^1)- G_{x^2_i}(x^1) )
- \sum_{i=1}^n  \ga_i (G_{x^1_i}(x^2)- G_{x^2_i}(x^2)) \\
& := g(x^1)- g(x^2),\end{array}$$ From the observation that
$K((x^1,x^2),(\hx^1,\hx^2)) =  \langle G_{x^1} -G_{x^2}, G_{\hx^1} -
G_{\hx^2}\rangle_G$, we also see that
$$\|\Im(g)\|_K = \|\sum_{i=1}^n  \ga_i K_{(x^1_i,x^2_i)}\|_K = \|\sum_{i=1}^n  \ga_i (G_{x^1_i}- G_{x^2_i})\|_G = \|g\|_G.$$ According to \cite{Aron}, the RKHS $\H_K$ is the completion of the above linear span of kernel sections $\{K_{(x^1_i,x^2_i)}:  x^1_i,x^2_i\in \X, i=1,\ldots,n \}$ and likewise, $\H_G$ is the completion of  the linear span of kernel sections $\{\{G_{x^i_1},G_{x^2_i}\}: x^1_i,x^2_i\in \X, i=1,\ldots,n \}$ which implies that, for any $f\in \H_K$, there exists $g\in \I_G$ such that $ f(x^1,x^2) =g(x^1)-g(x^2),$ and $\|f\|_K = \|g\|_G.$  It remains to prove that  $\Im(g)=0$ then $g\in \I_G.$ Indeed,  $\Im(g)(x^1,x^2)=0$ implies that $g(x^1) = g(x^2)$ for any $x^1,x^2\in \X.$ This means that $g$ is a constant function which means $g\in\I_G.$   This completes part (a) of the   proposition.

Part (b) follows from part (a) since, in this case, $\I_G = \{0\}$
which implies $\H_G = \I^\perp_G.$ This completes the proof of the
proposition.  \hfill $\Box$

Secondly, for the special pairwise kernel given by (\ref{eq:MLPK}),
we can establish the convergence of online pairwise learning
algorithm (\ref{eq:algorithm-2}) as stated in Theorem
\ref{exmthm:1}.

\noindent{\bf Proof of Theorem \ref{exmthm:1}:}  Part (a) directly
follows from Theorem \ref{thm:1}, Proposition \ref{prop-1} and the
definition of $\K_G$ given by (\ref{eq:K-G}).

For part (b), under the assumption $1_\X\not\in \H_G$, from
Proposition \ref{prop-1} we have
\begeqn\label{exmeqn-1}\beg{array}{ll}\K_G(s,\wtf_\rho)& \le
2\inf_{g\in \H_G} \{\|g-f_\rho\|_\rho+ {s\over 2} \|g\|_G\}\\ &\le
2\sqrt{2} \Bigl(\inf_{g\in\H_G}\{\|g-f_\rho\|^2_\rho + {s^2\over
4}\|g\|^2_G\}\Bigr)^{1/2}.\end{array}\endeqn According to
\cite{CS,Devito}, $\inf_{g\in\H_G}\{\|g-f_\rho\|^2_\rho + \gl
\|g\|^2_G\} \le  \gl^{2\gb} \|L^{-\gb}_G f_\rho\|_\rho$ for any
$\gb\le {1/2}.$  Now applying this estimation and (\ref{exmeqn-1}) with
$\gl = {s^2\over 4}$ and $s=\sum_{j=2}^t \gga_j$ implies that
$$\K_G(\sqrt{6}\gk(1+\gk)T^{-{1-\gth\over 2}},\wtf_\rho)  \le \O(T^{-(1-\gth)\gb}).$$
Putting this into (\ref{eq:converg}) and choosing $\gga_t = {1\over
\gk^2} t^{-{2\gb+1\over 2\gb+2}}$ yields the desired result. This
completes the proof of the theorem.  \hfill $\Box$

\section{Conclusion}\label{sec:conclusion}
This paper studied an online learning algorithm for
pairwise learning in an unconstrained RKHS setting called OPERA.   OPERA has a non-strongly convex objective function and is performed in an unconstrained setting, for which we are not aware
of similar studies for such online pairwise learning algorithms. We
established its almost-surely convergence  and
derived explicit error rates for polynomially decaying step sizes. Below we discuss some possible directions for
future work.

Firstly, the rates of OPERA under the regularity assumption
$\wtf_\rho \in L^\gb_K(L^2_\rho)$  are of the form $\EX[\|f_{T+1} -
\wtf_\rho\|_\rho] \le \O(T^{-{\gb\over 2\gb+2}}),$ which is
suboptimal compared with the rate $\O(T^{-{\gb\over 2\gb+1}})$
in the univariate case \cite{YP}.  It would be very interesting to
improve the rates of OPERA. Secondly, OPERA is not a fully online
learning algorithm since it needs to save previous samples $\bz^t =
\{(x_i,y_i): i=1,\ldots,t\}$ at iteration $t$, although, in the
linear case, efficient implementation may be possible. Hence, to improve the efficiency of OPERA the
other direction would be to introduce a memory-efficient
implementation which uses only a bounded subset of the past training samples as in \cite{Kar,Wang}. Finally, we only
considered the least-square loss for pairwise learning. It is
particularly interesting and challenging to establish similar
results for general loss functions.

\section*{Acknowledgements}

The work by D. X. Zhou described in this paper is supported by a
grant from the Research Grants Council of Hong Kong [Project No.
CityU 105011].

\section*{Appendix}

Here we present the proofs for Lemmas \ref{lemm:A} and Lemma
\ref{lemm:B}. To this end, we first state a technical lemma
which will be used later.
\beglemm\label{lemm:inequality-2} Let $\gga_j= {j^{-\gth}\over \mu}$ for any $j\in\N$
with $\gth\in (0,1)$. Then, for any $1\le j\le k$, there holds
\begeqn\label{eq:sum-poly}{1\over \mu(1-\gth)}((k+1)^{1-\gth} -
j^{1-\gth})\le \sum_{\ell=j}^k\gga_\ell \le {1\over
\mu(1-\gth)}(k^{1-\gth} - (j-1)^{1-\gth}).\endeqn
\endlemm
\noindent{\bf Proof.} Notice that $\ell^{-\gth} \le s^{-\gth}$ for $s\in
[\ell-1, \ell]$ and $\ell^{-\gth} \ge s^{-\gth}$ for $s\in
[\ell, \ell+1]$. Hence,
${1\over \mu}\sum_{\ell=j}^k\dint_{\ell}^{\ell+1} s^{-\gth}ds
\le \sum_{\ell=j}^k\gga_\ell \le {1\over \mu}
\sum_{\ell=j}^k\dint_{\ell-1}^\ell s^{-\gth} ds$ which implies
that
$${1\over \mu}\dint_{j}^{k+1} s^{-\gth}ds \le
\sum_{\ell=j}^k\gga_\ell \le {1\over \mu} \dint_{j-1}^k
s^{-\gth} ds.$$
The desired result follows directly from the above inequality.
\hfill $\Box$

We are ready to establish the proof of Lemma \ref{lemm:A}.

\noindent{\bf Proof of Lemma \ref{lemm:A}.} Let $\J :=
\sum_{j=2}^t {\gga_j (1+(\sum_{\ell=2}^{j-1}\gga_\ell)^{1/2}) \over
\sqrt{j}\bigl(1+ \sum_{\ell=j+1}^t \gga_\ell\bigr)^{1/2}}.$ It
can be written as
\begeqn\label{eq:J-decomp}\beg{array}{ll} \J & = \bigl[{\gga_t
(1+(\sum_{\ell=2}^{t-1}\gga_\ell)^{1/2}) \over \sqrt{t}} \bigr]+
\bigl[{\gga_2 \over \sqrt{2}\bigl(1+ \sum_{\ell=3}^t
\gga_\ell\bigr)^{1/2}} \bigr] + \bigl[ \sum_{j=3}^{t-1} {\gga_j
(1+(\sum_{\ell=2}^{j-1}\gga_\ell)^{1/2}) \over \sqrt{j}\bigl(1+
(\sum_{\ell=j+1}^t \gga_\ell)^{1/2}\bigr)^{1/2}}\bigr]\\
& := \J_1  + \J_2  + \J_3.\end{array}\endeqn
We estimate $\J_1,\J_2,$ and $\J_3$ separately as follows.

Firstly, let us look at the term $\J_1.$ Indeed, by
(\ref{eq:sum-poly}) we have \begeqn\label{eq:J1}\beg{array}{ll} \J_1
&= {1\over \mu}t^{-\gth-1/2} \bigl(1+{1\over
\sqrt{\mu}}(\sum_{\ell=1}^{t-1}\ell^{-\gth})^{1/2}\bigr)\le {1\over
\mu} t^{-\gth-1/2}\bigl(1+{t^{1-\gth\over 2}\over
\sqrt{\mu(1-\gth)}}\bigr)\\ & \le
{{2}\max(1,(\sqrt{\mu(1-\gth)})^{-1})} t^{-{3\gth\over
2}}.\end{array}\endeqn Secondly, for the term $\J_2$, we apply
(\ref{eq:sum-poly}) again to get that \begeqn\label{eq:J2}
\begin{array}{ll}\J_2 \le {2^{-\gth-1/2}\over \mu}
{\sqrt{\mu(1-\gth)}\over
\bigl((t+1)^{1-\gth}-3^{1-\gth}\bigr)^{1/2}} \le
\bigl({(1-\gth)\over (1-(3/5)^{1-\gth})\mu}\bigr)^{1/2}
t^{-(1-\gth)/2} \le \bigl({5\over 2\mu}\bigr)^{1/2}t^{-(1-\gth)/2},
\end{array}\endeqn where the second to last inequality used the
assumption $t\ge 4$ which implies $3^{1-\gth}\le ({3\over
5}(t+1))^{1-\gth},$, and the last inequality used the property that,
for any $0<\gth<1$ and $0<x<1$, that $(1-x)^{1-\gth}\ge
(1-\gth)(1-x).$

Lastly, we estimate the term $\J_3.$ To this end, by
(\ref{eq:sum-poly}) we can estimate $\J_3$ as follows:
\begeqn\label{eq:J3-1}\beg{array}{ll}\J_3 & \le {1\over \mu}
\sum_{j=3}^{t-1} {j^{-\gth} \bigl(1+{1\over
\sqrt{\mu(1-\gth)}}((j-1)^{1-\gth}-1)^{1/2}\bigr) \over \sqrt{j} \bigl(1+
{1\over \mu(1-\gth)}((t+1)^{1-\gth}-(j+1)^{1-\gth})\bigr)^{1/2}}
\\
& \le {{2}\over \mu}
\max(1,(\sqrt{\mu(1-\gth)})^{-1})\sum_{j=3}^{t-1} {j^{-{3\gth\over
2}} \over \bigl(1+ {1\over
\mu(1-\gth)}((t+1)^{1-\gth}-(j+1)^{1-\gth})\bigr)^{1/2}} \\
& \le {{2}\over \mu}
\max(\sqrt{\mu(1-\gth)})^{-1},\sqrt{\mu(1-\gth)}) \sum_{j=3}^{t-1}
{j^{-{3\gth\over 2}} \over \bigl(1+
((t+1)^{1-\gth}-(j+1)^{1-\gth})\bigr)^{1/2}}.
\end{array}\endeqn
It remains to estimate $\sum_{j=3}^{t-1} {j^{-{3\gth\over 2}} \over
\bigl(1+ ((t+1)^{1-\gth}-(j+1)^{1-\gth})\bigr)^{1/2}} .$ To this
end, we further decompose it into two terms as
\begeqn\label{eq:J3-2}\beg{array}{ll}\sum_{j=3}^{t-1}
{j^{-{3\gth\over 2}} \over (1+
((t+1)^{1-\gth}-(j+1)^{1-\gth}))^{1/2}} & = (\sum_{j>t/2}^{t-1} +
\sum_{3\le j\le t/2}) {j^{-{3\gth\over 2}}
\over (1+ ((t+1)^{1-\gth}-(j+1)^{1-\gth}))^{1/2}}\\
& := \wt{\J}_{31} + \wt{\J}_{32}.\end{array}\endeqn
For $\wt{\J}_{31}$, for any $s\in [j,j+1],$ that $j^{-\gth}\le
2^{\gth}(1+s)^{-\gth}$ and $(t+1)^{1-\gth} - (j+1)^{1-\gth} \ge
(t+1)^{1-\gth} - (s+1)^{1-\gth}.$ Then,
\begeqn\label{eq:J31}\beg{array}{ll} \wt{\J}_{31} & \le 2^{\gth\over 2}
t^{-{\gth\over 2}} \sum_{j>t/2}^{t-1} {j^{-\gth}\over (1+
((t+1)^{1-\gth}-(j+1)^{1-\gth}))^{1/2}} \\
& \le 2^{3\gth/2} t^{-{\gth\over 2}} \sum_{j>t/2}^{t-1}
\dint_j^{j+1} {(1+s)^{-\gth} ds\over
(1+(t+1)^{1-\gth}-(s+1)^{1-\gth})^{1/2}} \\
& \le  2^{3\gth/2} t^{-{\gth\over 2}} \dint_{t/2}^{t} {(1+s)^{-\gth}
ds\over (1+(t+1)^{1-\gth}-(s+1)^{1-\gth})^{1/2}}\\
&\le { 2^{1+3\gth/2} \over 1-\gth }t^{-{\gth\over
2}}\bigl[(1+(t+1)^{1-\gth}) -
(t/2+1)^{1-\gth}\bigr]^{1/2}\\
&\le {2^{1+3\gth/2}\over 1-\gth } t^{-{\gth\over 2}}
(t+1)^{1-\gth\over 2}\le {4 \sqrt{2} \over 1-\gth }t^{{1\over
2}-{\gth}} \end{array}\endeqn For $\wt{\J}_{32}$, the fact that
$(t+1)^{1-\gth}- (j+1)^{1-\gth} \ge (1-(2/3)^{1-\gth})
(t+1)^{1-\gth}$ for any $j\le t/2$ implies that
\begeqn\label{eq:J32} \wt{\J}_{32} \le {1\over t^{1-\gth \over 2}
(1-(2/3)^{1-\gth})} \sum_{3\le j<t/2} j^{-3\gth/2}\le {3\over
1-\gth}t^{-(1-\gth)/2}\sum_{3\le j<t/2} j^{-3\gth/2}. \endeqn Notice
that
$$\sum_{3\le j<t/2} j^{-3\gth\over 2} \le \int^{t/2}_2  s^{-3\gth/2}ds \le \left\{\beg{array}{ll}
{2\over |2-3\gth|}t^{-\min(0,{3\gth-2\over 2})}, & \hbox{ if } \gth\neq 2/3\\
  \ln t,  &  \hbox{ if } \gth=2/3
  \end{array}\right. $$
Putting the above inequality into (\ref{eq:J32}) yields that
\begeqn\label{eq:J32-1}\wt{\J}_{32} \le {A}_\gth
t^{-\min(\gth-{1\over 2}, {1-\gth\over 2})} \ln t,\endeqn where $
{A}_\gth = {6\over (1-\gth)|3\gth-2|}$ if $\gth\neq 2/3$ and
${3\over 1-\gth}$ otherwise. Combining (\ref{eq:J31}) and
(\ref{eq:J32-1}), (\ref{eq:J3-1}), and (\ref{eq:J3-2}) together
implies that \begeqn\label{eq:J3-full} \J_3 \le {B}_\gth
t^{-\min(\gth-{1\over 2}, {1-\gth\over 2})} \ln t, \endeqn where $$
{B}_\gth = \left\{\beg{array}{ll} {4 \max(\sqrt{\mu(1-\gth)})^{-1},
\sqrt{\mu(1-\gth)})\over \mu(1-\gth)} {(2\sqrt{2}+
{3\over |3\gth-2|})}, &  \hbox{ if } \gth\neq 2/3\\
{{2}(3+4\sqrt{2})\max(\sqrt{\mu(1-\gth)})^{-1}, \sqrt{\mu(1-\gth)})
\over \mu(1-\gth)}, & \hbox{ if } \gth=2/3.\end{array}\right.$$ Now
putting estimates (\ref{eq:J1}), (\ref{eq:J2}), and
(\ref{eq:J3-full}) together yields the desired result. This
completes the proof of the lemma. \hfill $\Box$

We now turn our attention to the proof for Lemma \ref{lemm:B}.

\noindent{\bf Proof of Lemma \ref{lemm:B}.} Let $\I =
\sum_{j=2}^t {\gga_j^2 (1+\sum_{\ell=2}^{j-1}\gga_\ell)
\over 1+ \sum_{\ell=j+1}^t \gga_\ell}.$ We can write $\I$ as
\begeqn\label{eq:I-decomp}\beg{array}{ll} \I & =
\bigl[{\gga_t^2} (1+\sum_{\ell=2}^{t-1}\gga_\ell) \bigr] +
\bigl[{\gga^2_2\over (1+\sum_{\ell=3}^{t-1}\gga_\ell) }\bigr]
+\bigl[\sum_{j=3}^{t-1} {\gga_j^2
(1+\sum_{\ell=2}^{j-1}\gga_\ell) \over 1+ \sum_{\ell=j+1}^t
\gga_\ell}\bigr]\\
& :=  \I_1 + \I_2 + \I_3,\end{array} \endeqn where we used the
conventional notation $\sum_{\ell=j+1}^{j} \gga_\ell =0$ for any
$j\in \N.$ We estimate $\I_1,\I_2,$ and $\I_3$ term by term as
follows.

Firstly, let us first estimate $\I_1$. By (\ref{eq:sum-poly}),
we can have that
\begeqn\label{eq:I1}\beg{array}{ll}\I_1 & \le {1\over \mu^2}
t^{-2\gth} (1+ {1\over \mu(1-\gth)}((t-1)^{1-\gth}-1))\\
& \le {2\max(1,(\mu(1-\gth))^{-1})\over \mu^2}
t^{1-3\gth}.\end{array}\endeqn

Secondly, we move on to the estimation of term $\I_2$. By
(\ref{eq:sum-poly}), we obtain that
\begeqn\label{eq:I2} \begin{array}{ll}\I_2 & \le {1\over 4\mu^2}
{1\over 1+{1\over \mu(1-\gth)}((t+1)^{1-\gth}-3^{1-\gth})}\\
& \le {1-\gth\over 4\mu\bigl(1-({3\over 5})^{1-\gth}\bigr)}
t^{-(1-\gth)}\le {5\over 8\mu} t^{-(1-\gth)}\end{array}\endeqn
where, in the second to last inequality, we used the assumption
$t\ge 4$ which implies $3^{1-\gth}\le ({3\over
5}(t+1))^{1-\gth},$ and the last inequality used the fact, for
any $0<\gth<1$ and $0<x<1$, that $(1-x)^{1-\gth}\ge
(1-\gth)(1-x).$

Finally, we turn our attention to the estimation of $\I_3.$
Applying (\ref{eq:sum-poly}) again to $\I_3$ implies that
\begeqn\label{eq:I3-1}\begin{array}{ll}\I_3 & \le {1\over \mu^2}
\sum_{j=3}^{t-1} {j^{-2\gth}\bigl(1+{1\over \mu(1-\gth)}
((j-1)^{1-\gth}-1)\bigr) \over 1+{1\over \mu (1-\gth)}
\bigl((t+1)^{1-\gth}-(j+1)^{1-\gth}\bigr)} \\
& \le {2\over \mu^2} \sum_{j=3}^{t-1} {j^{-2\gth}\max\big(1,{1\over
\mu(1-\gth)}\big) j^{1-\gth} \over 1+{1\over \mu (1-\gth)}
\big((t+1)^{1-\gth}-(j+1)^{1-\gth}\big)}\\
&\le {2\max(1,{(\mu(1-\gth)^{-1})})\max(1,\mu(1-\gth))\over
\mu^2}\sum_{j=3}^{t-1}
{j^{1-3\gth} \over 1+\big((t+1)^{1-\gth}-(j+1)^{1-\gth}\big)}\\
& \le {2\max(\mu(1-\gth),{(\mu(1-\gth)^{-1})})\over
\mu^2}\sum_{j=3}^{t-1} {j^{1-3\gth} \over
1+\big((t+1)^{1-\gth}-(j+1)^{1-\gth}\big)}.\end{array}\endeqn It now
suffices to estimate the term $\I_3:=\sum_{j=3}^{t-1} {j^{1-3\gth}
\over 1+\big((t+1)^{1-\gth}-(j+1)^{1-\gth}\big)},$
 which can be written as
\begeqn\label{eq:I3-2}\beg{array}{ll}\I_3 & =
\bigl(\sum_{j>t/2}^{t}+ \sum_{3\le j\le t/2}\bigr)
{j^{1-3\gth} \over 1+\big((t+1)^{1-\gth}-(j+1)^{1-\gth}\big)}\\
&: = \wt{\I}_{31} + \wt{\I}_{32}. \end{array}\endeqn
For the first term $\wt{\I}_{31},$ observe, for any $s\in [j,j+1],$ that
$j^{-\gth}\le 2^{\gth}(1+s)^{-\gth}$ and $(t+1)^{1-\gth} -
(j+1)^{1-\gth} \ge (t+1)^{1-\gth} - (s+1)^{1-\gth}.$ Therefore,
\begeqn\label{eq:I3-3}\beg{array}{ll}\wt{\I}_{31} & : =
\sum_{j>t/2}^{t-1} {j^{1-3\gth} \over
1+\big((t+1)^{1-\gth}-(j+1)^{1-\gth}\big)} \\
& \le 2^{2\gth-1} t^{1-2\gth} \sum_{j>t/2}^{t-1}
\dint_{j}^{j+1}{(s+1)^{-\gth} \over
1+\big((t+1)^{1-\gth}-(s+1)^{1-\gth}\big)} ds \\
& \le 2^{2\gth-1}t^{1-2\gth} \dint_{t/2}^t {(s+1)^{-\gth} \over
1+\big((t+1)^{1-\gth}-(s+1)^{1-\gth}\big)} ds \\
& ={2^{2\gth-1}t^{1-2\gth}\over 1-\gth} \bigl[ \ln
(1+\big((t+1)^{1-\gth}-(t/2)^{1-\gth}\big)) - \ln
(1+\big((t+1)^{1-\gth}-t^{1-\gth}\big))\bigr] \\
& \le {2^{2\gth-1}t^{1-2\gth}\over 1-\gth}  \ln (t+1)^{1-\gth} \le
2^{2\gth-1} t^{1-2\gth} \ln (t+1)\le 4 t^{1-2\gth} \ln t.
\end{array}\endeqn
For $\wt{\I}_{32}$, we have
\begeqn\label{eq:I3-4}\beg{array}{ll}\wt{\I}_{32} & = \dsum_{3\le
j\le t/2} {j^{1-3\gth} \over
1+\big((t+1)^{1-\gth}-(j+1)^{1-\gth}\big)} \\
& \le \dsum_{3\le j\le t/2} {j^{1-3\gth} \over
1+(1-({2/3})^{1-\gth})(t+1)^{1-\gth}}\\
&\le {t^{-(1-\gth)}\over (1-({2/3})^{1-\gth})}\dsum_{3\le j\le
t/2} {j^{1-3\gth}}\le {3t^{-(1-\gth)}\over (1-\gth)}\dsum_{3\le
j\le t/2} {j^{1-3\gth}},\end{array}\endeqn
where we used again the fact, for any $0<\gth<1$ and $0<x<1$,
that $(1-x)^{1-\gth}\ge (1-\gth)(1-x).$
Also, by a simple calculation, there holds
$$ \dsum_{3\le j\le t/2} {j^{1-3\gth}} \le
\left\{\beg{array}{ll} {1\over |2-3\gth|} t^{-\min(0,3\gth-2)}, &
\hbox{ if }
\gth\neq 2/3\\
\ln t ,  &  \hbox{ if } \gth = 2/3.\end{array}\right.$$ Putting the
above estimation into (\ref{eq:I3-4}) yields that
\begeqn\label{eq:I3-5} \wt{\I}_{32} \le \widetilde{A}_\gth \;
t^{-\min(2\gth-1,1-\gth)}\ln t. \endeqn where $\widetilde{A}_\gth =
{3\over |3\gth-2| (1-\gth)} $ if $\gth\neq 2/3$ and ${3\over
1-\gth}$ otherwise. Putting (\ref{eq:I3-3}) and (\ref{eq:I3-5}) back
into (\ref{eq:I3-1}) implies that \begeqn\label{eq:I3} \wt{\I}_{3}
\le \widetilde{B}_\gth \; t^{-\min(2\gth-1,1-\gth)}\ln t,\endeqn
where $\widetilde{B}_\gth ={3\over (1-\gth)|3\gth-2|}+4$ if
$\gth\neq 2/3$ and ${3\over 1-\gth} + 4$ otherwise. Combining
estimates (\ref{eq:I1}), (\ref{eq:I2}), and (\ref{eq:I3}) together
yields the desired result. This completes the proof of the lemma.
\hfill $\Box$

\end{document}